\documentclass{article}

\usepackage{arxiv}
\usepackage{multirow}
\usepackage[utf8]{inputenc}
\usepackage[T1]{fontenc}   
\usepackage{hyperref}      
\usepackage{url}           
\usepackage{booktabs}      
\usepackage{amsfonts}
\usepackage{amsmath}      
\usepackage{nicefrac}      
\usepackage{microtype}     
\usepackage{cleveref}      
\usepackage{graphicx}
\usepackage[square, numbers, sort&compress]{natbib}
\usepackage{doi}
\usepackage{caption}
\usepackage{subcaption}
\usepackage{bm}
\usepackage{algorithm}
\usepackage{algpseudocode}
\usepackage{setspace}
\usepackage[inkscapepath=svgsubdir]{svg}
\usepackage{tipa}
\usepackage{listings}
\usepackage{subcaption}
\usepackage{xcolor}
\usepackage{makecell}
\usepackage{float}
\usepackage{lineno}

\definecolor{codegreen}{rgb}{0,0.6,0}
\definecolor{codegray}{rgb}{0.5,0.5,0.5}
\definecolor{codepurple}{rgb}{0.58,0,0.82}
\definecolor{backcolour}{rgb}{0.95,0.95,0.92}

\lstdefinestyle{mystyle}{
    backgroundcolor=\color{backcolour},   
    commentstyle=\color{codegreen},
    keywordstyle=\color{magenta},
    numberstyle=\tiny\color{codegray},
    stringstyle=\color{codepurple},
    basicstyle=\ttfamily\scriptsize,
    breakatwhitespace=false,         
    breaklines=true,                 
    keepspaces=true,                 
    numbersep=5pt,                  
    showspaces=false,                
    showstringspaces=false,
    showtabs=false,                  
    tabsize=2
}
\newcommand{\fw}{0.8\linewidth}
\newcommand*{\revcolorflag}{black} % red for review 1 else black
\newcommand*{\rev}[1]{\textcolor{\revcolorflag}{#1}}
\newcommand*{\rrevcolorflag}{black} % red for review 2 else black
\newcommand*{\rrev}[1]{\textcolor{\rrevcolorflag}{#1}}

\graphicspath{{./figures}}

\title{Towards real-time surrogate-free Bayesian inversion for neutron reflectometry}
\renewcommand{\shorttitle}{Towards real-time surrogate-free Bayesian inversion for neutron reflectometry}

\date{}
\author{
	\href{https://orcid.org/0000-0002-3037-7584}{\includegraphics[scale=0.06]{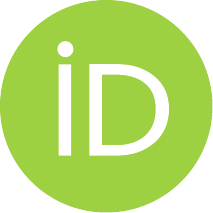}\hspace{1mm}Max D.~Champneys}
	\thanks{Corresponding author} \\
	Dynamics Research Group\\
	University of Sheffield\\
	Sheffield, UK\\
	\texttt{max.champneys@sheffield.ac.uk} \\
	\And
	\href{https://orcid.org/0000-0001-8606-8644}{\includegraphics[scale=0.06]{orcid.pdf}\hspace{1mm}Andrew J.~Parnell } \\
	School of Mathematical and Physical Sciences\\
	University of Sheffield\\
	Sheffield, UK\\
	\texttt{a.j.parnell@sheffield.ac.uk} \\
	\And
	\href{https://orcid.org/0000-0002-7412-8571}{\includegraphics[scale=0.06]{orcid.pdf}\hspace{1mm}Philipp Gutfreund} \\
	Institut Laue-Langevin\\
	Grenoble, France\\
	\texttt{gutfreund@ill.eu} \\
	\And
\href{https://orcid.org/0000-0003-0086-2965}{\includegraphics[scale=0.06]{orcid.pdf}\hspace{1mm}Maximilian W. A. Skoda } \\
	ISIS Pulsed Neutron and Muon Source\\
	Rutherford Appleton Laboratory\\
	Didcot, UK\\
	\texttt{maximilian.skoda@stfc.ac.uk} \\
	\And
	\href{https://orcid.org/0000-0002-1675-5219}{\includegraphics[scale=0.06]{orcid.pdf}\hspace{1mm}J. Patrick A. Fairclough} \\
	School of Mechanical, Aerospace and Civil Engineering\\
	University of Sheffield\\
	Sheffield, UK\\
	\texttt{p.fairclough@sheffield.ac.uk} \\
	\And
\href{https://orcid.org/0000-0002-3433-3247}{\includegraphics[scale=0.06]{orcid.pdf}\hspace{1mm}Timothy J.~Rogers} \\
	Dynamics Research Group \\
	University of Sheffield \\
	Sheffield, UK\\
	\texttt{tim.rogers@sheffield.ac.uk} \\
	\And
\href{https://orcid.org/0000-0001-7870-9843}{\includegraphics[scale=0.06]{orcid.pdf}\hspace{1mm}Stephanie L.~Burg} \\
	School of Chemical, Material and Biological Engineering\\
	University of Sheffield \\
	Sheffield, UK\\
	\texttt{s.l.burg@sheffield.ac.uk} \\
}

\hypersetup{
	pdftitle={\shorttitle},
	pdfsubject={Neutron Reflectometry, Automatic differentiation},
	pdfauthor={Max D Champneys},
	pdfkeywords={},
}

\begin{document}

\maketitle

\begin{abstract}
\small
Neutron reflectometry (NR) is a key enabling technology for many areas of scientific development. Although the forward reflectivity model is well-known, inferring the physical properties of a sample from NR data requires the solution of an inverse problem.  Increasingly, beamline scientists are using NR in fast kinetic configurations and probing highly-complex structures and interfaces, introducing significant uncertainty. Existing uncertainty quantification (UQ) approaches in NR, such as Markov-Chain Monte-Carlo (MCMC), suffer from poor sample efficiency and slow convergence times. Recently, surrogate machine learning models have been proposed as an alternative. However, physical intuition is lost when replacing governing equations with fast surrogates. Instead, we propose a rapid, surrogate-free Bayesian inversion for NR. Our approach offers a step-change in inference speed and efficiency. For the first time in NR, exact gradients through the reflectivity are computed, enabling highly performant gradient-based inference schemes: Hamiltonian Monte-Carlo offers significant advances in sample efficiency compared to MCMC. Variational inference enables approximate UQ on the order of seconds rather than hours. We demonstrate state-of-the-art performance on a thick oxide quartz film, and robust co-fitting performance in the high complexity regime of organic LED multilayer devices. Additionally, we provide an open-source library of reflectometry kernels in the python language.

\end{abstract}

\keywords{Neutron Reflectometry \and Bayesian Inference \and Automatic differentiation}

\section{Introduction}

Neutron reflectometry (NR) is a powerful technique to probe surfaces and interfaces. As NR is inherently an indirect measurement technique, access to the physical quantities of interest (layer thickness, scattering length density, roughness), necessitate the solution of an inverse modelling problem, that is inefficient for large amounts of data or complex multilayer structures (e.g. lithium batteries / electrodes). This inefficiency is particularly pronounced when \emph{uncertainty quantification} is desired. Sampling schemes such as Markov-chain Monte-Carlo (MCMC) can suffer from slow convergence and poor sample efficiency as the number of unknown parameters increases. Recently, surrogate machine learning models have been proposed as an alternative to existing optimisation and inference routines. Although such approaches have been successful in some areas, physical intuition is invariably lost when replacing governing equations with fast neural networks or other surrogates.

In this paper, we seek to address the shortcomings of existing approaches to uncertainty quantification (UQ) in NR. The key innovation is that we access the gradients of the forward specular reflectivity of a thin multilayer film w.r.t the unknown parameters of interest, unlocking a host of powerful optimisation and inference techniques that remain thus-far unexploited in the context of neutron reflectometry. The principal advantages of such an approach are:

\begin{itemize}
	\item Order of magnitude gains in sample-efficiency (using Hamiltonian Monte-Carlo) and inference speed (using a variational approach) compared to gradient-free MCMC approaches.
	\item The exact forward reflectivity laws are used to perform the inversion, no approximate pre-trained surrogate model is required.    
	\item Full Bayesian posterior that can accommodate arbitrary prior knowledge.
\end{itemize}

The proposed approach is demonstrated in two NR case study examples: A benchmark Quartz film and a co-fit to four LED devices. Additionally, an open-source software library is provided for fast computation of forward reflectivity and gradients in the python language at \url{github.com/MDCHAMP/refjax}.

\section{Background}

NR is a widely used technique for the nanometer scale characterisation of soft matter \cite{skoda2019recent}, thin films \cite{kirby2012phase} and bio-materials \cite{junghans2015analysis}. Increasingly, this has involved the study of more complex layered structures and ones that are evolving in time (dynamic). The application of NR has lead to many high-profile breakthroughs in the physical and biological sciences. Polarized neutron reflectivity (PNR) was crucial for understanding the oscillatory exchange coupling behaviour so fundamental to giant magnetoresistance (GMR) \cite{Majkrzak1988}, a pivotal modern technology. Ongoing work has investigated how surfactant mixtures perform at varying temperatures using surface tension and neutron reflectivity \cite{liley2019performance}, this has been crucial for developing more efficient, low-temperature detergents, vital for reducing the carbon footprint of washing laundry. Ongoing NR studies have also been important for the study of biological membranes \cite{pabst2010applications}, their interaction with drugs \cite{clifton2015accurate}, such as antimicrobial peptides and other antibiotics.

NR is a scattering technique that illuminates a sample of interest using thermal neutrons produced and moderated from either a reactor or a spallation neutron source. NR measures the specular reflectivity $R$ as a function of the momentum transfer $Q$ from counting neutrons using a detector \cite{penfold1990application}. Theoretically, the forward reflectivity model of complex materials is well understood. Given the physical and apparatus parameters $\theta$, one can appeal to Fresnel reflectivity to compute the predicted specular reflectivity $\hat{R}$. NR is however, inherently an indirect technique, which requires the reflectivity data to be modelled, as it is the absolute square of Fourier transform of the scattering length density, which gives rise to a loss in phase information. The parameters of interest cannot be directly accessed from measurements of the reflectivity. Fitting a parametric model of the sample is the inverse problem at the core of NR measurements and data analysis.

In the four decades since the widespread adoption of NR as a measurement technique, many methods have been proposed to solve the inverse modelling problem. Until recently, advances in NR analysis and fitting have kept close pace with the progress in machine learning (ML). The rapid rise of deep learning, automatic differentiation and specialised computation libraries has revolutionised the scalability of ML in many other fields. Yet, these analysis routes have so far been very slow to translate to NR fitting. Among the leading software packages for NR and thin-film analysis \cite{nelson2019refnx, woollam2017completeease}, the underlying fitting routines are often based on traditional data optimisation techniques such as Levenberg-Marquardt (LM) \cite{levenberg1944method, marquardt1963algorithm}, Broyden-Fletcher-Goldfarb-Shanno (BFGS) \cite{nocedal2006numerical}, or heuristic techniques such as differential evolution \cite{storn1997differential} or particle swarms \cite{eberhart1995new}. Other approaches follow Bayesian methodologies such as Markov-chain Monte-Carlo (MCMC) \cite{metropolis1953equation, hastings1970monte} or nested sampling \cite{skilling2004nested}. The authors are not currently aware of any widely-used NR software that fully utilises the power of modern automatic-differentiation and gradient-based optimisation techniques to directly solve the inverse problem.

In place of physically derived models, surrogate approaches have very recently been proposed to replace the inverse problem \cite{doucet2021machine, hinderhofer2023machine, pithan2023closing, starostin2025fast, rentzsch2026towards} with ML. These ML models supplant the forward reflectivity model and are trained to predict the parameters of interest directly from the measured $R$. Although such approaches are able to leverage advanced ML technologies, it is inevitable that physical intuition is lost when replacing governing equations with fast machine learning.

\subsection{The NR inverse problem}

% \begin{figure*}
% 	\centering
% 	\includegraphics[width=\fw]{schematic_new.png}
% 	\caption{The inverse problem at the heart of NR. Both the sample and the reflectometer instrument are described by unknown parameters $\theta$. By selection of an appropriate error function $e$, the identification of the unknown parameters is reduced to an optimisation problem.}
% 	\label{fig:schematic}
% \end{figure*}

At the very core of NR data analysis, is the inverse problem of obtaining estimates of unknown parameters of interest pertaining to both the slab model of the sample $\theta_{\text{slab}}$ (layer thicknesses, scattering length densities and roughnesses) and the instrument $\theta_\text{NR}$ (scaling/alignment parameters and the background detection level). For notational simplicity, in the description that follows these unknown parameters shall be referred to jointly as $\theta = \{\theta_{\text{NR}}, \theta_{\text{slab}}\}$. The Abeles matrix formalism \cite{abeles1949theorie} provides a physical model of the reflectivity $\hat{R}(Q, \theta)$ in terms of these parameters and the momentum transfer. Predictions from the forward model can be compared to the observed reflectivity in order to determine the values of the unknown parameters. A key challenge in NR, is the non-uniqueness of the inverse problem because the phase information is lost when measuring the reflected neutron intensity. For a particular measured neutron reflectivity measurement $R(Q)$, there may be many possible slab model solutions and \rev{therefore many sets of parameters that describe the data equally well. In practice, there are a number of ways to mitigate the lack of phase information. One approach (as discussed in \cite{kirby2012phase}) is to use multiple independent datasets to perform the inversion. Another approach uses reference layers with known density profiles.}

\rev{In the absence of additional measurements, optimisation routines can be initialised in the vicinity of optima that are physically plausible given prior knowledge of the sample under investigation. In this work we will consider the problem of inversion in the vicinity of physically plausible solutions. Such local optimisation is an area in which gradient-based methods excel, particularly in high-dimensions.}

Methods for solving the NR inverse problem can be broadly categorised into two groups that will be discussed in this section. The first group comprises \emph{Optimisation methods}, these seek to find an optimal set of parameters $\theta^*$ that optimally satisfy an error metric $e$,

\begin{equation}
	\theta^* | R =  \underset{\theta}{\mathrm{argmin}} \  e\left(R(Q, \theta), \hat{R}(Q, \theta)\right)
\end{equation}

Alternatively, \emph{Inference methods} seek to describe the unknown parameters probabilistically, accounting for various sources of uncertainty in the analysis. Such approaches commonly apply Bayes rule to enumerate the posterior distribution,

\begin{equation}
	p(\theta | R) = \frac{p(\theta)p(R | \theta)}{p(R)}
\end{equation}

\noindent where $p(\theta)$ is the prior over parameters, $p(R | \theta)$ is the likelihood model that encompasses both the forward reflection model and the instrument parameters and $p(R)$ is the (usually intractable) marginal data likelihood. Since the observed $R(Q)$ is derived from count data, it may be possible to derive an exact likelihood function. However, the authors note that in practice it is often sufficient to approximate $p(R | \theta)$ as a Gaussian with variances derived from the count data directly.

\begin{figure*}
	\centering
	\includegraphics[width=\fw]{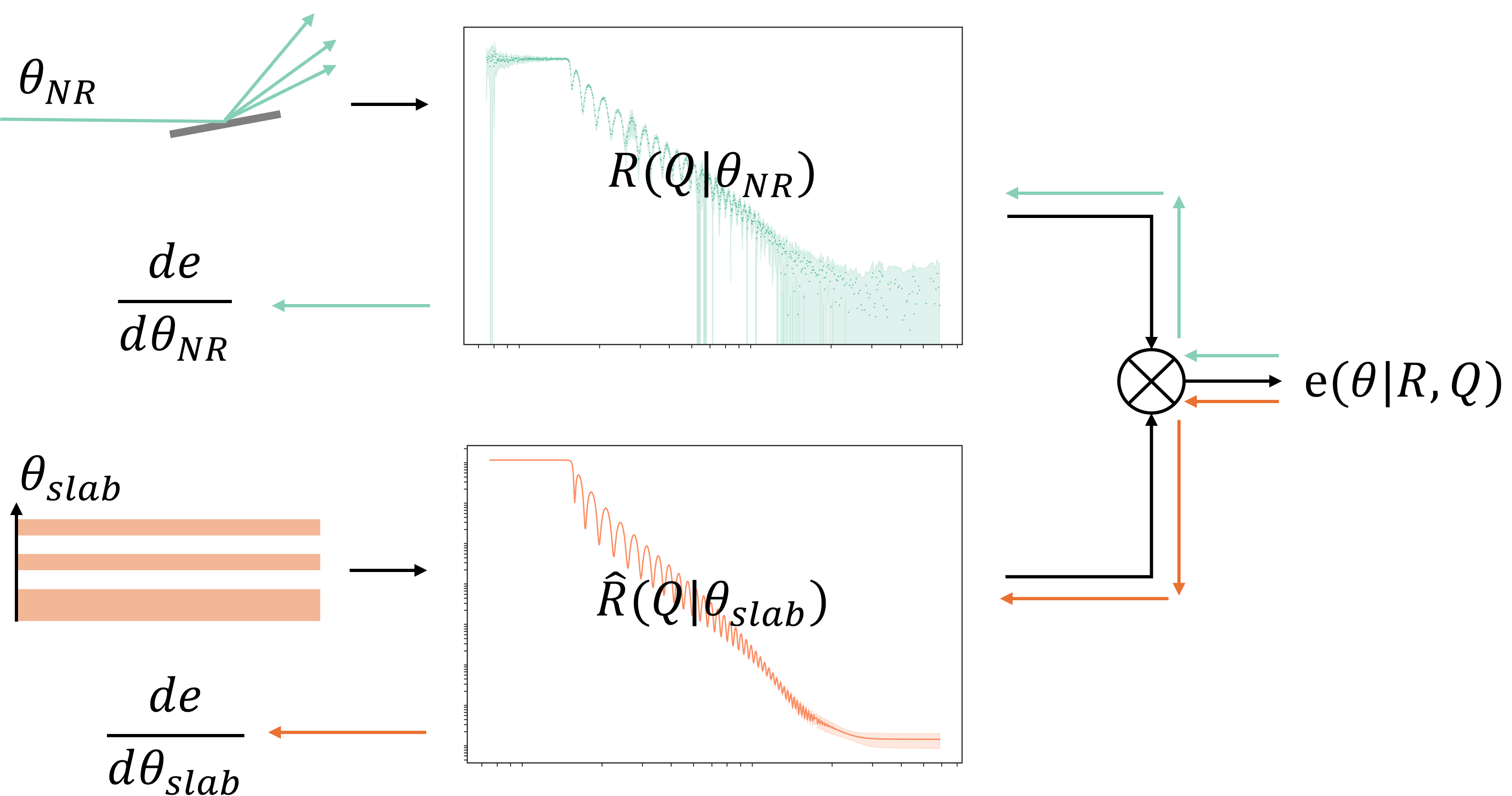}
	\caption{Automatic-differentiation back propagates the gradient of the error function through the forward reflectivity model by successive application of the chain rule. The result is the gradient of the error function with respect to the parameters of interest.}
	\label{fig:schematic_grad}
\end{figure*}

In both the optimisation and inference setting, powerful techniques based on automatic differentiation exist in the ML literature for solving inverse problems. Access to the gradients of the error function (from this work, Figure \ref{fig:schematic_grad}) means that these approaches can be readily applied to NR. The remainder of this section will be dedicated to introducing approaches from both categories that are applied in the case studies outlined in this paper.

\subsection{Optimisation for NR by gradient descent}
\label{sec:opt_NR}

Although several approaches that use approximate gradient information \cite{levenberg1944method, marquardt1963algorithm, nocedal2006numerical} \rev{are common in the NR literature, we are not aware of any method that explicitly harnesses} the analytical gradient with respect to the parameters of interest using automatic differentiation techniques. Methods in this class have seen intensive development since the rise of deep neural networks, particularly stochastic gradient descent methods. Access to gradient information as in this paper will make the application of these schemes very straightforward to the NR community. Efficient implementations of many popular algorithms are readily available in a number of programming languages. The most popular methods (e.g. Adam \cite{kingma2014adam}, RMSPROP \cite{hinton2012rmsprop}) make use of gradient averages between iterations and momentum terms that help to overcome local minima. Such approaches help to alleviate some of the inherent shortcomings of these popular gradient descent approaches when applied to NR data analysis, such as becoming trapped in local minima and computational efficiency in the case of high dimensions of very many parameters.

\subsection{Inference for NR by Hamiltonian Monte-Carlo}

In the interest of keeping this paper self-contained and in order to appeal to practitioners of NR who may not be familiar with gradient-based sampling and approximate inference schemes, \rev{brief overviews of Hamiltonian Monte-Carlo (HMC) and variational inference (VI) for NR practitioners are included here}. The interested reader is directed to \cite{betancourt2017conceptual} (HMC) and \cite{blei2017variational} (VI) for more comprehensive tutorial overviews of the method. 

The essential idea behind HMC is to construct a pseudo-dynamical system in the space of the unknown NR and slab parameters, designed such that the stationary density of the resulting position vector is the target posterior distribution of interest,

\begin{equation}
	\theta(t) \sim p(\theta| R)
\end{equation}

Thus, one can simply simulate the dynamic system and interpret the time series as samples from the posterior in the same way as a Markov-chain approach. By requiring that the proposed dynamics of $\theta(t)$ are Hamiltonian, one is able to express the joint distribution of position $\theta$ and momentum $r$ in terms of the potential and kinetic energies.

\begin{equation}
	H(\theta, r) = U(\theta) + K(r)
\end{equation}

\begin{equation}
	p(\theta, r) \propto \exp(-H(\theta, r)) = \exp(-U(\theta)) \exp(-K(r))
\end{equation}

From the above, it can be straightforwardly shown that the choice $U(\theta)= - \log p(\theta| R)$ leads to the required stationary density for $\theta$. A common choice for the momentum distribution is to choose a zero-mean Gaussian distribution with `mass' matrix $M$ as a hyperparameter of the method $K(r) = \frac{1}{2} r^T M^{-1} r$. The resulting Hamiltonian system is governed by the ordinary differential equation (ODE),

\begin{equation}
	\begin{aligned}
		\frac{d\theta}{dt} & = \frac{\partial H}{\partial r} = M^{-1}r                                                                                  \\
		\frac{dr}{dt}      & = -\frac{\partial H}{\partial \theta} = -\nabla_\theta U(\theta) = \nabla_\theta \log \left[p(\theta) p(R | \theta)\right]
	\end{aligned}
\end{equation}

Where the differentiation has removed the unknown normalisation constant in the kinetic energy and the ODE now depends on the hyperparameters, likelihood and prior only. As with any other nonlinear ODE it is necessary to integrate forward in time numerically in order to obtain trajectories (\rev{and therefore samples from the target posterior distribution over the slab and NR parameters}). Symplectic integrators are a class of integrator specifically designed to simulate Hamiltonian dynamics, by attempting to conserve (or bound the drift in) the systems total energy through the Hamiltonian. A popular solver in the context of HMC is the leapfrog algorithm \cite{hairer2006geometric}. For a given integration period $\Delta t$, the dynamics are updated sequentially as,

\begin{equation}
	\begin{aligned}
		r_{t + \frac{1}{2}} & = r_t - \frac{\Delta t}{2} \nabla_\theta U(\theta_t)                     \\
		\theta_{t+1}        & = \theta_t + \Delta t \, M^{-1} r_{t + \frac{1}{2}}                      \\
		r_{t+1}             & = r_{t + \frac{1}{2}} - \frac{\Delta t}{2} \nabla_\theta U(\theta_{t+1})
	\end{aligned}
\end{equation}

Thus, the overall sampling scheme consists of a Markov chain with the following kernel. In each iteration, an initial momentum is drawn, and the integration is run for a number of steps $L$ with a step size of $\Delta t$. Although the position at the end of the integration window should theoretically be a valid sample from the target distribution, in practice, numerical errors in the integration can lead to bias. To overcome this, a Metropolis-Hastings step is typically taken after integration. At the end of the integration window, the last value of the parameter is accepted as the next state of the Markov Chain with probability $\alpha$. It can be shown (in \cite{betancourt2017conceptual} for example) that the correct choice of $\alpha$ to ensure detailed balance in the Markov Chain is given by,

\begin{equation}
	\alpha = \min \Big\{1, \exp\left(H(\theta, r) - H(\theta', r')\right)\Big\}
\end{equation}

\rev{\subsubsection{HMC for NR practitioners}}

\rev{In practice, the performance of HMC comes to depend strongly on the choice of the hyperparameters $L, \Delta t$ and $M$. The optimal setting of these parameters strongly affects the sampling efficiency (how many samples must be drawn in order to robustly explore the posterior), and is affected by the geometry of the posterior, such as by parameters that vary on vastly different scales. For NR practitioners working with HMC, it is advisable to pre-scale parameters such that the parameter variances have the same order of magnitude. An example would be fitting the thickness of a wide layer, or a very small background term subject to a fixed scaling value to improve numerical conditioning.}   

\rev{A further factor that can lower sample efficiency in HMC is sharp curvature in the posterior geometry. A typical example in NR is parameters with hard constraints e.g. layer thicknesses or SLDs constrained to be positive. Close to a zero value, the posterior has infinite curvature and thus the leapfrog integrator presented above can quickly accumulate errors. A common mitigation that can work well in the context of NR inversion is to run the HMC sampler in log space and then exponentiate for the physical parameters once the sampling is completed. }

\rev{Fortunately, the adaptive specification of HMC hyperparameters is a well studied problem. Variants of the standard HMC algorithm such as the no U-turn sampler (NUTS) \cite{hoffman2014no} have shown incredible robustness to a wide range of posterior geometries. NUTS is now a standard approach in Bayesian inference and implementations can be found in many programming languages. }

\rev{With regard to evaluation of the performance of HMC in the context of NR, all methods that are appropriate for MCMC can be applied to HMC. A common statistic is the \emph{effective sample size} \cite{gelman2013bayesian}, a measure of representation power that is lower for chains that are non-stationary and poorly converged. The authors note that although acceptance rate can be a useful measurement of sampler performance, in HMC it is not unusual to see rates above 80\% in well converged chains.  }

\subsection{Approximate inference for NR by variational inference}

HMC has shown good performance across a range of inference scenarios. However, in the limit of very large data and many unknown parameters, even advanced sampling schemes can suffer from slow convergence and high computational complexity. In such settings, variational approaches (collectively known as variation inference (VI)) have shown significant promise by approximating the true posterior with a parametric distribution \cite{jordan1999introduction}. VI essentially transforms the inference problem back into an optimisation problem that can be solved by gradient descent methods.

Variational inference methods differ from sampling approaches in that they attempt to directly \emph{approximate} the posterior distribution with some parametric form that is straightforward to sample from,

\begin{equation}
	p(\theta|R) \approx q(\theta ; \phi)
\end{equation}

Where $\phi$ are some deterministic unknown parameters that control the shape of a surrogate distribution $q$. It is important here that the choice of $q$ is a valid distribution (that it integrates to unity and covers the support of the posterior) regardless of the parameters $\phi$. For this reason it is common to choose families of probabilities distributions or mixture models for $q$ \cite{hoffman2013stochastic, yin2018semi}.

With $q$ selected, all that remains is to find the optimal setting of $\phi$ \rev{such that the surrogate distribution forms the best possible approximation to the true (but unknown) posterior distribution $p(\theta | R)$ over the unknown slab and NR parameters.} To achieve this, we first require a notion of how far our approximation is from the true posterior. To measure divergence between probability distributions, we can use the Kullback-Leibler divergence (KL divergence), defined as,

\begin{equation}
	KL(q||p) = \mathbb{E}_q\left[\log \frac{q(\theta;\phi)}{p(\theta|R)}\right]
\end{equation}

thus, the optimal choice for the variational parameters $\phi$ is,

\begin{equation}
	\phi^* = \operatorname*{arg\,min}_{\phi} KL(q(\theta;\phi)||p(\theta|R))
\end{equation}

Note that this form still contains the unknown posterior $p(\theta | R)$, and so some rearrangement is required. Recalling that from Bayes rule,

\begin{equation}
	p(\theta | R) = \frac{p(\theta)p(R|\theta)}{p(R)}
\end{equation}

\begin{align}
	KL(q||p) & = \mathbb{E}_q\left[\log \frac{q(\theta;\phi)p(R)}{p(\theta)p(R|\theta)}\right]       \\
	         & =\mathbb{E}_q\left[\log q(\theta;\phi) - \log{p(\theta)p(R|\theta)} +\log p(R)\right]
\end{align}

Since the log evidence is constant we can bring it outside the expectation,

\begin{equation}
	KL(q||p) = \mathbb{E}_q\left[\log q(\theta;\phi) - \log{p(\theta)p(R|\theta)} \right] +\log p(R)
\end{equation}

Given that we want to minimise the KL divergence between $q$ and $p$ and that it is invariant to changes in $\phi$ we can essentially ignore this term in the optimisation and focus on the expectation in the above. This expectation term is often called the `evidence lower bound' (ELBO) \rrev{\cite{blei2017variational}} because it provides a lower bound on the unknown evidence term. This can be proved with Jensen's inequality for probability distributions \cite{jordan1999introduction}. For a concave function $f$,

\begin{equation}
	f(\mathbb{E}[\theta]) \geq \mathbb{E}[f(\theta)]
\end{equation}

Now rewriting the log evidence term as,

\begin{align}
	\log p(R) & = \log \int p(\theta)p(R|\theta) d\theta                                       \\
	          & = \log \int p(\theta)p(R|\theta) \frac{q(\theta;\phi)}{q(\theta;\phi)} d\theta \\
	          & = \log \mathbb{E}_q\left[ \frac{p(\theta)p(R|\theta)}{q(\theta;\phi)}\right]
\end{align}

since the logarithm is a concave function, Jensen's inequality gives that,

\begin{equation}
	\log p(R) = \log \mathbb{E}_q\left[ \frac{p(\theta)p(R|\theta)}{q(\theta;\phi)}\right] \geq \underbrace{\mathbb{E}_q\left[ \log \frac{p(\theta)p(R|\theta)}{q(\theta;\phi)}\right]}_{\text{ELBO}}
\end{equation}

Therefore, maximising the ELBO is equivalent to minimising the KL divergence,

\begin{equation}
	\phi^* = \operatorname*{arg\,max}_{\phi} \text{ELBO}
\end{equation}

Thus, the appropriate loss function is,

\begin{equation}
	\mathcal{L}(\phi) = - \text{ELBO} = \mathbb{E}_q\left[\log q(\theta|\phi) - \log{p(\theta)p(R|\theta)} \right]
\end{equation}

Note that these expectations are rarely available analytically. In practice, they are approximated by Monte-Carlo. A common choice in the VI literature is to make this approximation using a single sample \cite{kingma2014auto}.

\begin{equation}
	\mathcal{L}(\phi)\approx \frac{1}{N} \sum^N_{i=1} \left[\log q(\theta^{(i)}|\phi) - \log{p(\theta^{(i)})p(R|\theta^{(i)})} \right]
\end{equation}

where the $\theta^{(i)}$ are the $N$ samples drawn from $q$ to approximate the expectation. Taking the gradient of the above with respect to the unknown parameters $\phi$,

\begin{equation}
	\nabla_\phi\mathcal{L}(\phi)\approx \frac{1}{N} \sum^N_{i=1} \nabla_\phi \left[\log q(\theta^{(i)}|\phi) -  \log{p(\theta^{(i)})p(R|\theta^{(i)})} \right]
\end{equation}

It has been observed empirically that single sample Monte-Carlo approximations to the ELBO have low variance and give good performance \cite{kingma2014auto}. However, care must be taken to ensure that the gradients remain available through the sampling procedure. A common approach is to reparametrise $q$ such that the sampling procedure does not depend on $\phi$\footnote{For more detail on the reparameterisation trick see \cite{kingma2014auto}.}. 

\rev{\subsubsection{VI for NR practitioners}}

\rev{In the context of NR, the loss gradient above once again requires access to the gradient of the forward reflectivity model, made possible by the contribution of this work. With access to the gradient of the ELBO, all that remains is to solve the optimisation problem with a gradient descent scheme such as one of those described in section \ref{sec:opt_NR}. }

\rev{It is important to note here that VI based on minimisation of the KL divergence can display `mode-seeking' behaviour that emphasises high probability regions of the posterior at the expense of lower probability areas leading to underprediction of variance. For a discussion of why this occurs and advice in the context of NR see \rrev{Supplementary Information Section~S3}. }

\rev{In the limit of very large data (>10k samples), computing the expectations above (even with a single sample approximation) may become prohibitively expensive. In such contexts, a particularly powerful extension is \emph{Stochastic variational inference} (SVI) \cite{hoffman2013stochastic}. \rev{The results of this paper indicate that SVI is particularly promising in the large data, complex model regime of NR having been successfully applied to datasets numbering millions of examples in other fields \cite{hoffman2013stochastic}.} However, because most static NR datasets comprise <1k samples, only a standard VI approach is considered in this work.}

\rev{An important choice in VI is the selection of a surrogate distribution $q$ that will approximate the true posterior. In an NR context it is important that the support of this distribution covers the expected support of the parameters of interest. For example, if two parameters are expected to be correlated (e.g. thickness and SLD in a kinetic heating experiment) then a $q$ must be selected that can represent this correlation. For example, independent Gaussian distributions would be unable to reproduce this correlation, whereas a multivariate Gaussian would (at the cost of additional parameters to be optimised).
}
\subsection{Related work}

The authors are not aware of previous contributions directly applying automatic differentiation to the forward reflectivity model to perform NR inversion in the literature\footnote{\rev{Although no direct applications of automatic-differentiation for NR inversion are to be found in the literature, the authors are aware of at least two library implementations that support this functionality in addition to our own contribution. The jax backend in refnx \cite{nelson2019refnx} which supports automatic differentiation, although this functionality is not exposed to the user through the user interface and differentiation with respect to refnx model parameters is not possible due to architectural constraints. Very recently, the reflectorch package \cite{Munteanu2025} has been published with a pytorch backend that naturally supports automatic differentiation and tensor computation. This library has been used to generate training examples of NR data for neural networks in \cite{munteanu2024neural}.}}. However, recent significant interest has been shown in simulation-based surrogate modelling in both the optimisation and inference settings.

In several works, neural-network surrogates have been proposed as an alternative to optimisation for the NR inverse problem in order to extract unknown parameters directly from data. Methods have been proposed in both NR \cite{doucet2021machine, rentzsch2026towards} and X-ray diffraction \cite{pithan2023closing}, with the latter motivated as a way to provide real-time process control based on surrogate inversions. A recent review of ML technology applied to scattering problems can be found in \cite{hinderhofer2023machine}. In such cases the principal idea is to pre-train a neural-network surrogate on indicative reflectometry curves (either simulated or experimental) such that it is able to directly extract the parameters of interest. The key advantage is that for a new unseen testing reflectometry curve, the parameters can be extracted in `one-shot' without the need for potentially computationally expensive algorithms when solving the NR inverse problem. \rev{While such approaches have merit, it is apparent that they are weakened by the requirement to train the methods on large quantities of appropriately labelled training data} (experimental or otherwise). While there are certainly situations in which this will be plausible, \rrev{such methods require training data representative of the sample under investigation, which can be difficult to guarantee when the structure is unknown.}

Recently, surrogate modelling has been proposed in an inference setting for NR. In \cite{starostin2025fast}, Starostin et al. propose to apply techniques from neural posterior estimation and prior amortisation to train a neural network to produce credible prior distributions based on reflectivity curves. The idea is that the network is trained to rapidly identify credible regions of the parameter space in the form of prior distributions that can be supplied to traditional Bayesian inference techniques such as Markov-chain Monte-Carlo (MCMC) or importance sampling (IS). The result is a significant reduction in the cost of evaluating a posterior over the NR parameters. The results of the paper are impressive and show that credible prior distributions can be found that include the multimodality that occurs due to the phase invariance problem when using very wide priors. As with all simulation based modelling approaches however, the requirement to pre-train a model (in this case on simulated data from a fixed number of layers) remains a limitation of the approach.

\section{refjax - an open-source library for gradient based NR}

Rather than implement any single bespoke gradient-based optimisation or inference technique for NR, the authors present here an open-source software library for fast computation of forward reflectivity and gradients in the python language. The library, titled refjax, is available at \url{github.com/MDCHAMP/refjax}. The library is built on top of the python JAX ecosystem from Google \cite{jax2018github} that provides flexible automatic differentiation, parallelism and GPU acceleration. refjax provides simple access to the forward specular reflectivity calculation (including smearing) for an arbitrary slab model. In refjax, all input parameters are collected in standard python dictionaries with a standardised format.

\begin{lstlisting}[language=Python]

theta = {
    "thick": jnp.array([    1500, 10.0     ]), # thicknesses (n,)   n layers (A)
    "SLD":   jnp.array([0.0, 5.0, 4.2, 2.07]), # SLD         (n+2,) fronting + n layers + backing (A^-2)
    "rough": jnp.array([   5.0, 5.0, 2.07  ]), # roughnesses (n+1,) interfaces between media
    "scale": 1.0,
    "bkg":   2e-7,
}
\end{lstlisting}

These parameters can then be provided to a kernel to compute the forward reflectivity.

\begin{lstlisting}[language=Python]
import refjax
kernel = refjax.kernel(refjax.model.abeles) 
R_hat = kernel(theta, Q)
\end{lstlisting}

Because the library is built upon the JAX ecosystem, just in time compilation, parallelism and gradient computation are handled naturally.

\begin{lstlisting}[language=Python]
kernel_jit = jax.jit(kernel) # just-in-time compilation

def mean_squared_error(theta): # define a loss function
	R_hat = kernel_jit(theta, Q)
	return ((R-R_hat)**2).sum()

de_dtheta_fn = jax.grad(mean_squared_error)(theta) # returns gradient of loss

jax.pmap(kernel_jit)(thetas, Qs) # parallel evaluations of the kernel+gradient are possible on GPU/TPU etc.
\end{lstlisting}

The module also provides a number of smearing kernels including constant and pointwise Gaussian smearing.

\begin{lstlisting}[language=Python]
kernel = refjax.kernel(abeles, refjax.smear.constant_Gaussian)
smeared_R_hat = kernel_pointwise(theta, Q, dQ)
\end{lstlisting}

A key advantage of the provided library is the freedom to apply any optimisation routine developed in the JAX ecosystem and beyond. To date, highly performant packages are available for traditional optimisers (LM, BFGS) \cite{rader2024optimistix}, (stochastic) gradient descent \cite{deepmind2020jax}, HMC (including the no U-turn sampler \cite{hoffman2014no}) \cite{phan2019composable} and others. Several examples of how to use the library in conjunction with popular ML optimisation suites are provided at \url{https://github.com/MDCHAMP/refjax/tree/main/examples}.

\section{Case studies}

The effectiveness of gradient-based approaches to NR inverse problems and the software we have developed is demonstrated here for two case-study examples. In each case the forward reflectivity computation and automatic differentiation was implemented using the refjax software.

\subsection{Crystalline quartz}

In the first case study, the performance of gradient based techniques are highlighted on an existing dataset. To this end, the parameters of a highly homogeneous crystal quartz film deposited on a flat silicon wafer with a thin native oxide layer are considered. The reflectivity data were collected on the D17 neutron reflectivity instrument at the ILL (Grenoble, France) \cite{gutfreund2018towards}. As part of the data collection, pointwise full-width at half maximum (FWHM) smearing kernel estimates were estimated. These were used to perform smearing in the forward reflectivity model as fixed known parameters.
The slab model considered in this case study is provided in Table \ref{tbl:quartz} and the NR parameters are displayed in Table \ref{tbl:quartz_nrparam}. All parameters are considered unknown and are fit using both the optimisation and inference methods.

\renewcommand{\arraystretch}{1.5}
\begin{table}
	\centering
	\caption{The Slab model for the Quartz sample including initial parameter guesses for optimisation, priors used for inference and identified posterior mean and standard deviations. Note the large range in prior values corresponding to very weak prior knowledge.}
	\label{tbl:quartz}
	\begin{tabular}{lllll}
		Layer                                 & Parameter                & \makecell{Initial                                                     \\ value} & \makecell{Prior\\ distribution}                        & \makecell{Posterior\\(mean $\pm$ stdev.)} \\\hline\hline
		\multirow{2}{*}{Air}                  & SLD ($\text{\AA}^{-2}$)  & 0.0               &                               &                   \\
		                                      & Thickness ($\text{\AA}$) & $\infty$          &                               &                   \\\hline
		                                      & \rev{Interfacial roughness ($\text{\AA}$)}    & 5.0               & $\mathcal{U}(2.0, 20.0)$      & $11.402\pm0.0827$ \\\hline
		\multirow{2}{*}{Quartz (Si02)}        & SLD ($\text{\AA}^{-2}$)  & 5.0               & $\mathcal{U}(0.0, 5.0)$       & $4.240\pm0.0015$  \\
		                                      & Thickness ($\text{\AA}$) & 1500              & $\mathcal{U}(1400.0, 1500.0)$ & $1470\pm0.487 $   \\\hline
		                                      & \rev{Interfacial roughness ($\text{\AA}$)}    & 5.0               & $\mathcal{U}(2.0, 20.0)$      & $10.905\pm0.43$   \\\hline
		\multirow{2}{*}{Native oxide (Si02)}  & SLD ($\text{\AA}^{-2}$)  & 4.2               & $\mathcal{U}(0.0, 5.0)$       & $0.323\pm0.216$   \\
		                                      & Thickness ($\text{\AA}$) & 10                & $\mathcal{U}(0.0, 50.0)$      & $2.122\pm0.415$   \\\hline
		                                      & \rev{Interfacial roughness ($\text{\AA}$)}    & 5.0               & $\mathcal{U}(2.0, 20.0)$      & $3.163\pm0.733$   \\\hline
		\multirow{2}{*}{Silicon backing (Si)} & SLD ($\text{\AA}^{-2}$)  & 2.07              & $\mathcal{U}(0.0, 5.0)$       & $2.195\pm0.0236$  \\
		                                      & Thickness ($\text{\AA}$) & $\infty$          &                               &                   \\
	\end{tabular}
\end{table}

\begin{table}
	\centering
	\caption{NR parameters for the quartz data including initial parameter guesses for optimisation, priors used for inference and identified posterior mean and standard deviations.}
	\label{tbl:quartz_nrparam}
	\begin{tabular}{llll}
		Parameter       & Initial value & Prior distribution        & Posterior ($\pm \sigma$)       \\\hline\hline
		Background      & $10^{-7}$     & $\mathcal{U}(e^{-20}, 1)$ & $(1.05\pm 0.201)\times10^{-7}$ \\
		Scale parameter & 1.0           & $\mathcal{U}(0.9, 1.5)$   & $1.048\pm0.00363$              \\
	\end{tabular}
\end{table}

\begin{figure}
	\centering
	\includegraphics[width=\fw]{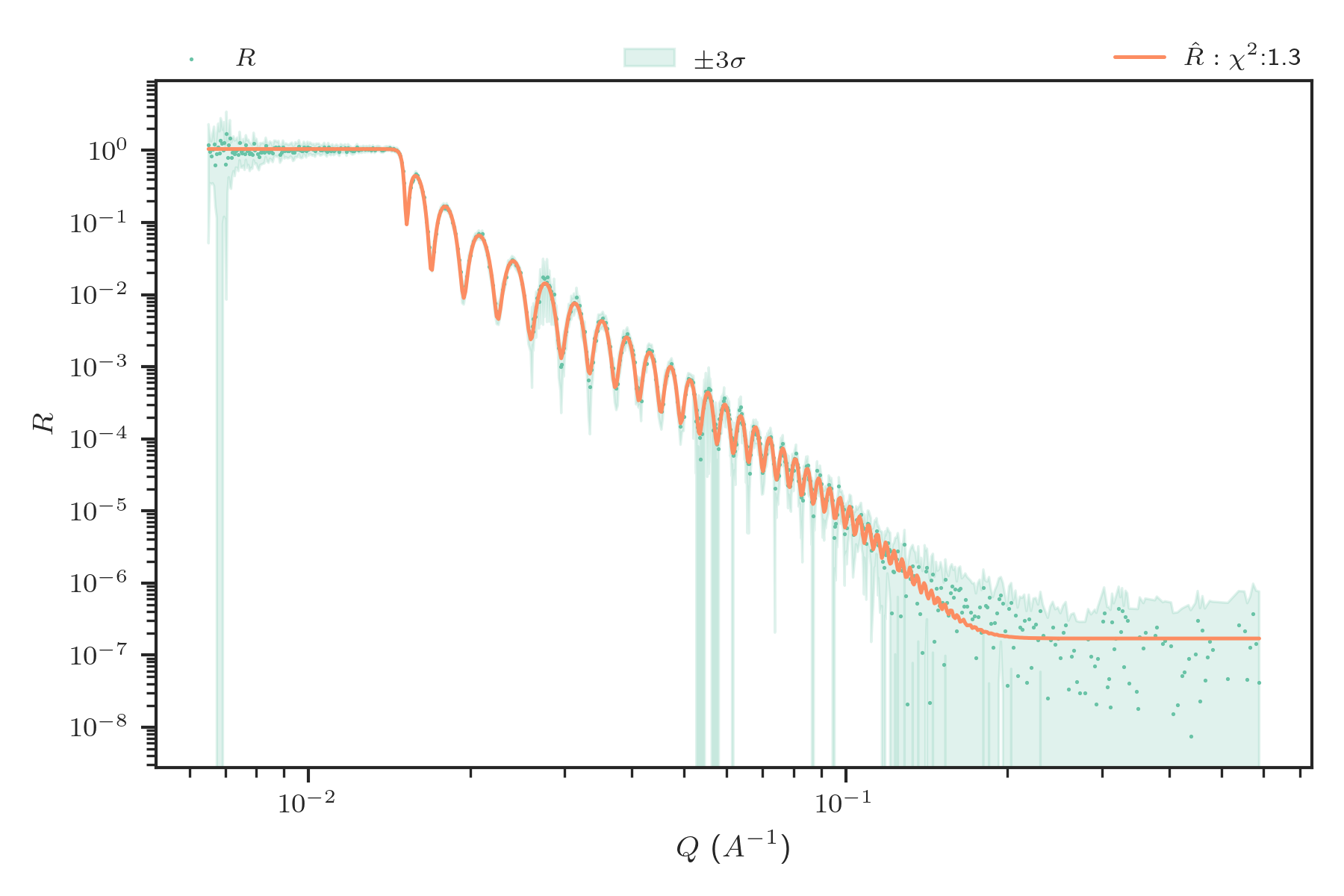}
	\caption{Optimisation of quartz thin film NR dataset fit using gradient decent with the ADAM optimiser \cite{kingma2014adam}. All parameters are floating and so free to fit. The identified $\chi^2$ error is lower than the value of $\chi^2=1.32$ reported previously in \cite{gutfreund2018towards} for the same data.}
	\label{fig:q_adam}
\end{figure}

First, an optimisation approach is considered. To perform the model fit, the widely-used stochastic gradient descent technique ADAM \cite{kingma2014adam} is employed. Starting from an approximate initial values for the relevant parameters (listed in Tables \ref{tbl:quartz} and \ref{tbl:quartz_nrparam}), the optimiser descends the gradient of a $\chi^2$ error metric,

\begin{equation}
	\chi^2 = \sum \frac{(\hat{R}(Q) - R(Q))^2}{\sigma_{R(Q)}^2}
\end{equation}

for 2000 iterations with a learning rate of 0.05. The resulting fit to the data is given in Figure \ref{fig:q_adam}. As can be seen in the Figure, the fit to the data is excellent, achieving a $\chi^2$ error of 1.302. This is marginally lower than the value of $\chi^2=1.32$ reported in \cite{gutfreund2018towards}, \rev{indicating that the approach has found optimal parameters. Note that because the optimisation is unconstrained, and the data are not very sensitive to the natural oxide layer, the fitted values for this layer should be interpreted with care.} 

With the effectiveness of gradient based optimisation established, a further study considering gradient-powered inference for NR fitting is presented here. The same quartz data is used to perform Bayesian inference of the unknown parameters by HMC. The NUTS sampler \cite{hoffman2014no} (as implemented in the `numpyro' library \cite{phan2019composable}) is used to sample from the posterior distribution over the parameters. NUTS has the advantage that the hyperparameter $L$ is set automatically in each iteration leading to efficient exploration of the typical set. The other HMC parameters are adapted using the default strategy in the numpyro library for a burn-in period of 1000 samples.

\begin{figure}
	\centering
	\includegraphics[width=\fw]{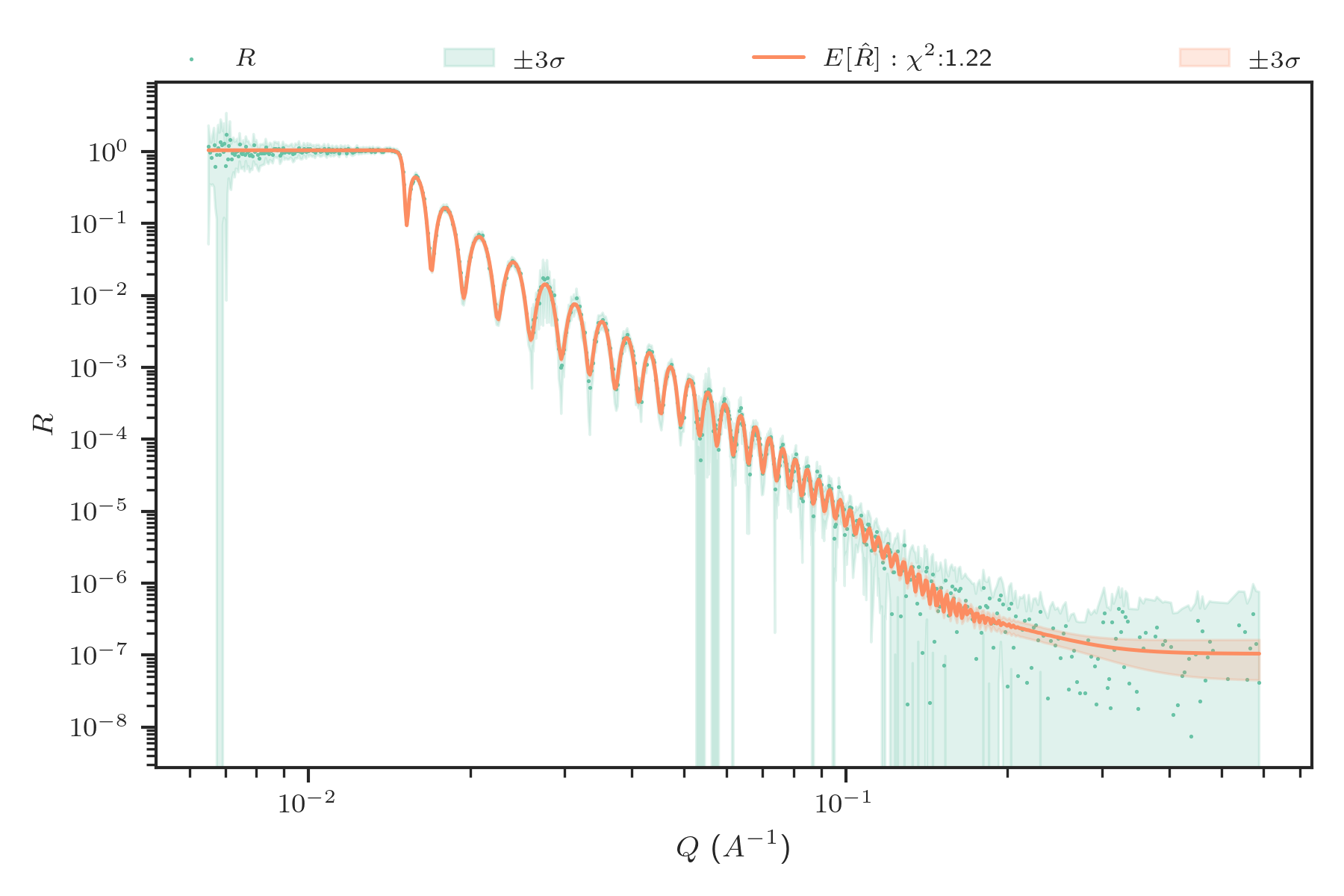}
	\caption{Inference of quartz NR model by Hamiltonian Monte-Carlo. All parameters remain free to fit. It is notable that the mean of the posterior distribution has a lower error than values identified by the deterministic optimisation routine. Posterior uncertainty is given by the width of the orange shaded region, corresponding to a $\pm 3\sigma$ interval.}
	\label{fig:q_nuts}
\end{figure}

\begin{figure}
	\centering
	\includegraphics[width=\fw]{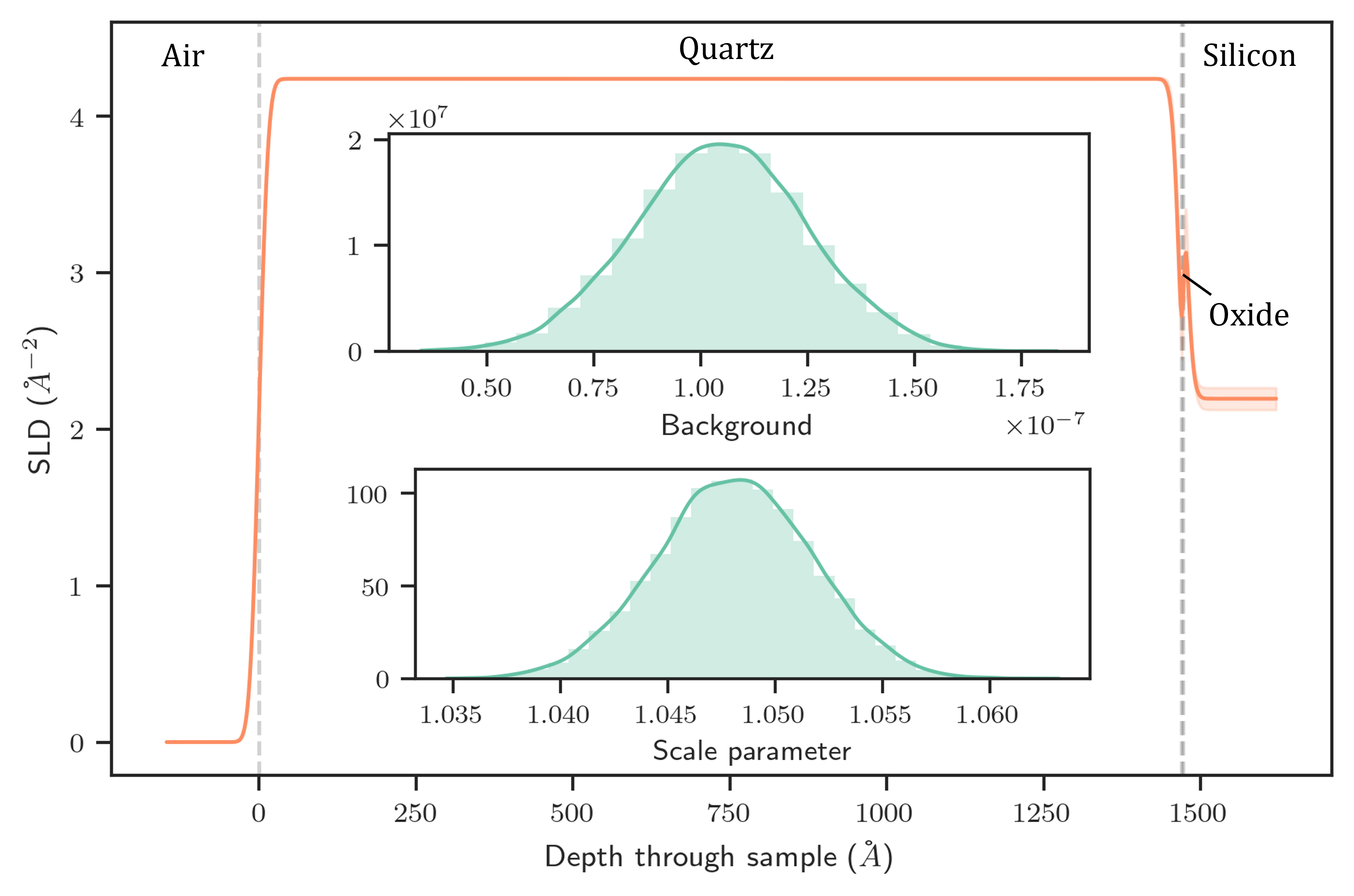}
	\caption{Posterior SLD profile and posterior marginal parameter distributions for the NR parameters of the quartz film, obtained by Hamiltonian Monte-Carlo. Error bars on the SLD profile represent $\pm3\sigma$ intervals, these are visible on the right for the oxide and Silicon.}
	\label{fig:q_inf}
\end{figure}

The prior distributions were set to be uniform in the vicinity of the initial parameters used by the gradient scheme (see Tables \ref{tbl:quartz} and \ref{tbl:quartz_nrparam}). \rev{Note that the Bayesian approach enables us to incorporate prior knowledge and constrain these parameters to physically-plausible values.} The chain was initialised prior to the burn-in at the optimal value found by gradient descent. Overall, 10 independent chains were run, each comprising 2000 samples drawn after burn-in. Posterior means and standard deviations are recorded in Tables \ref{tbl:quartz} and \ref{tbl:quartz_nrparam}. The resulting predictive posterior $p(\hat{R}|R)$ is plotted in Figure \ref{fig:q_nuts}, wherein an excellent fit to the data is seen\footnote{\rev{Taking the posterior mean parameters and performing further optimisation using a BFGS optimiser (SciPy implementation) results in a $\chi^2$ score reduction of only 0.004 indicating that the inference has found optimal parameters.}}. The mean prediction of the posterior $E[\hat{R}]$ has a $\chi^2$ error of 1.22, an improvement over previously published results ($\chi^2=1.32$, \cite{gutfreund2018towards}) although we note that in our work, all parameters are left free to fit. In Figure \ref{fig:q_inf}, the marginal posterior distributions of the NR parameters of interest are shown alongside the distribution over the SLD profile of the quartz film. % the 2 vs 3 layer question - wait for reply from Philip/Andy - not a huge issue

It is noteworthy that the obtained results are obtained by drawing only 2000 samples from the NUTS algorithm per chain, a fraction of the number of samples typically required by MCMC methods for good convergence. This phenomenon is in line with the efficiency gains expected from HMC over traditional MCMC methods, as demonstrated in other fields \cite{hajian2007efficient, tinto2018gravitational}. However, it is of interest to quantify this phenomenon in the context of NR inversion.

\begin{figure}
	\centering
	\includegraphics[width=\fw]{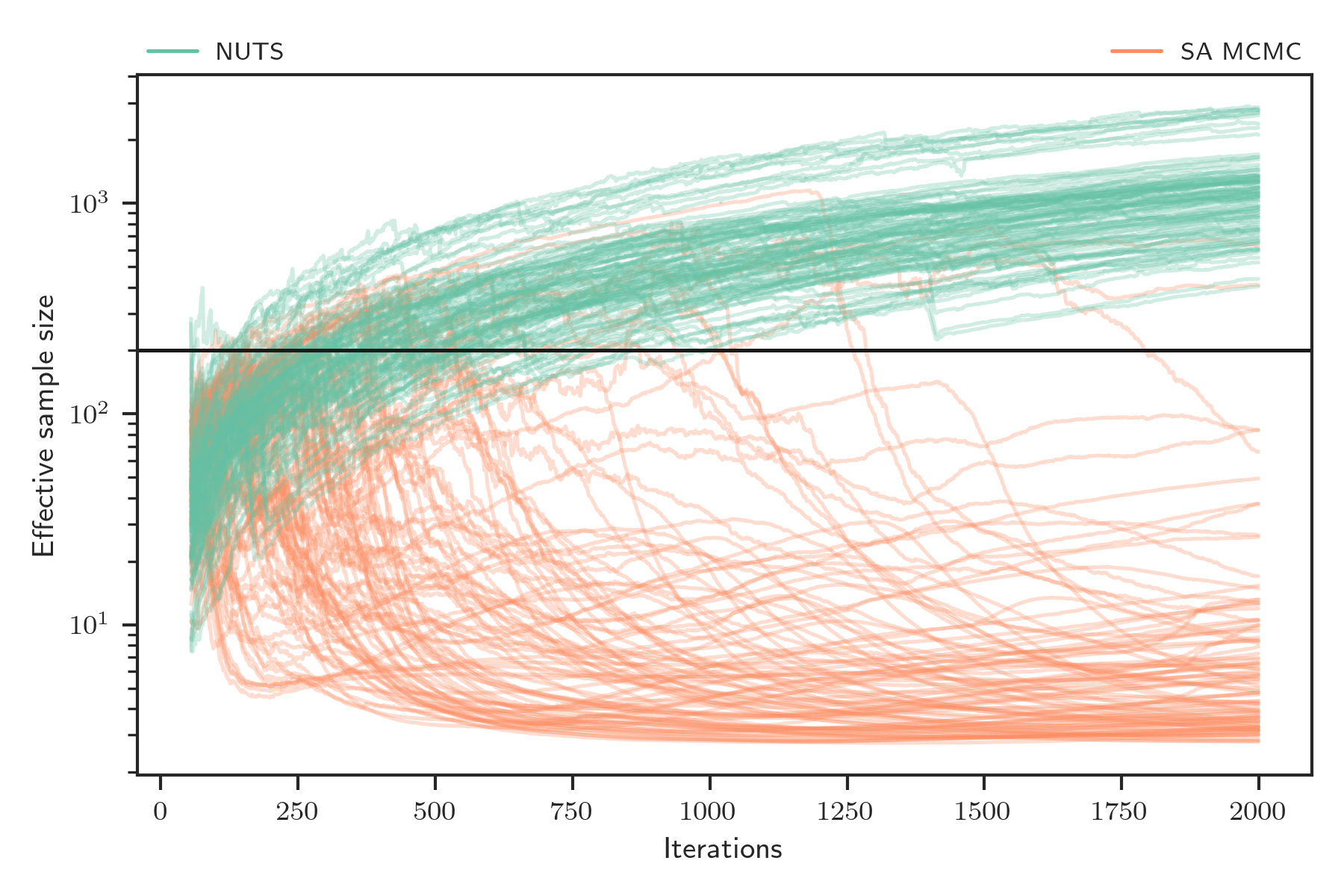}
	\caption{Comparison of effective sample sizes per iteration between a Hamiltonian Monte-Carlo (NUTS) scheme (green) and a gradient-free sample adaptive MCMC approach (orange). Each line is the effective sample size of a single parameter in a single chain. The horizontal line corresponds to a sample size of 200 which is a common rule of thumb for well converged chains in the literature \cite{lanfear2016estimating}.}
	\label{fig:ES}
\end{figure}

In order to compare the efficiency of HMC (here NUTS) against gradient-free MCMC methods, Figure \ref{fig:ES} plots the effective sample size (ESS) against iteration for both HMC and a Sample-adaptive MCMC (SA-MCMC) approach \cite{zhu2019sample} implemented in the numpyro library \cite{phan2019composable}. ESS gives a measure of how many independent samples a correlated Markov-chain is equivalent to. A common rule of thumb in the Bayesian inference literature is that an ESS value of 200 per parameter represents a good coverage of the posterior for well-converged chains, whereas low values can correspond to poor exploration and autocorrelation.

Samples from both NUTS and SA-MCMC are collected from 10 independent chains for 2000 iterations, starting from the same initialisation point at the optimal parameter values found by gradient descent. At each iteration the ESS for each parameter in each chain is plotted in Figure \ref{fig:ES}.

As can be seen in Figure \ref{fig:ES}, the NUTS samples increase in ESS broadly monotonically with iteration, indicating good convergence and efficient exploration of the typical set even after only 1000 samples. SA-MCMC appears to start well (by exploring a local region of the typical set efficiently) but slow mixing and high auto-correlation lead to a decrease in ESS. It is interesting to note that although the per-sample computational complexity for NUTS is several times higher than SA-MCMC, better exploration of the typical set is seen with the gradient-based scheme, in line with results from other fields \cite{hajian2007efficient, tinto2018gravitational}.

\subsection{Organic LED}

With good performance shown on data from a relatively simple slab model, it is next of interest to consider how gradient-based approaches perform on more complex samples. Here, data from several organic light-emitting diode devices architectures are presented that correspond to different stages in the processing of these devices.

The organic  (OLED) devices are fabricated using spin coating. They comprise three main layers. The first being Poly(3,4-ethylenedioxythiophene):poly(styrene sulfonate) (PEDOT:PSS): This serves as the hole injection layer (HIL), facilitating charge transfer from the anode. Then Poly[(9,9-dioctylfluorenyl-2,7-diyl)-co-(4,4'-(N-(4-sec-butylphenyl)diphenylamine))] (TFB): This layer functions as both a hole transport layer (HTL) and an electron blocking layer (EBL), ensuring efficient charge recombination in the emissive zone. Finally, Poly[(9,9-di-n-octylfluorenyl-2,7-diyl)-alt-(benzo[2,1,3]thiadiazol-4,8-diyl)] (F8BT), this is the emissive layer, responsible for green light emission upon charge recombination. The TFB was deuterated to provide suitable neutron scattering length density contrast to accurately probe the interfacial mixing of TFB and F8BT.

A slab model of the LED device is given in Table \ref{tbl:LED}. Overall, four devices are considered. The first is not annealed after manufacture and was tested as cast. The other three were annealed at temperatures of 80ºC, 140ºC and 180ºC respectively. Because smearing kernel sizes are not available for this dataset, a constant Gaussian smear is adopted with its size left as a further unknown parameter (see Table \ref{tbl:LED_nrparam}). \rrev{The four devices were measured on the same instrument and were fabricated on nominally identical silicon substrates. It is therefore physically reasonable to treat only the instrument-related parameters (the background level and the resolution smearing) and the substrate scattering length density as being shared between devices. We emphasise that the physically meaningful slab parameters of the polymer layers (their thicknesses, SLDs and interfacial roughnesses) are left entirely free and are inferred independently for each device. The co-fit therefore does not assume that the samples are identical; it encodes only the fact that they share a common measurement apparatus and substrate, which reduces the number of instrument nuisance parameters that must be inferred.} For this reason (and to demonstrate the effectiveness of variational inference for NR) the parameters of all four devices are inferred jointly, with the background and smearing parameters being common to all four devices. In addition, the SLD value of the silicon backing (which is expected to be very close to the literature value of 2.07 for all four devices) is treated as a co-fit parameter. In general, wider priors are adopted that apply to all layers, representing weak prior knowledge. However, in order to encourage physically plausible roughness values around the native silicon oxide layer, the prior over the roughnesses either side of the oxide are constrained to values less than 8.0. In total this results in a posterior distribution over 59 unknown parameters. Given the large parameter size, approximate inference with VI will be applied here.

\begin{table}
	\centering
	\caption{The Slab model for the LED data including initial parameter guesses for optimisation and priors used for inference.}
	\label{tbl:LED}
	\begin{tabular}{l|lll}
		Layer                                 & Parameter                & \makecell{Initial                             \\ value} & \makecell{Prior\\ distribution}                      \\\hline\hline
		\multirow{2}{*}{Air}                  & SLD ($\text{\AA}^{-2}$)  & 0.0               &                           \\
		                                      & Thickness ($\text{\AA}$) & $\infty$          &                           \\\hline
		                                      & \rev{Interfacial roughness ($\text{\AA}$)}    & 10.0              & $\mathcal{U}(0.0, 30.0)$  \\\hline
		\multirow{2}{*}{F8BT}                 & SLD ($\text{\AA}^{-2}$)  & 1.0               & $\mathcal{U}(0.0, 5.0)$   \\
		                                      & Thickness ($\text{\AA}$) & 700.0             & $\mathcal{U}(0.0, 800.0)$ \\\hline
		                                      & \rev{Interfacial roughness ($\text{\AA}$)}    & 10.0              & $\mathcal{U}(0.0, 30.0)$  \\\hline
		\multirow{2}{*}{TFB}                  & SLD ($\text{\AA}^{-2}$)  & 4.0               & $\mathcal{U}(0.0, 5.0)$   \\
		                                      & Thickness ($\text{\AA}$) & 400.0             & $\mathcal{U}(0.0, 800.0)$ \\\hline
		                                      & \rev{Interfacial roughness ($\text{\AA}$)}    & 10.0              & $\mathcal{U}(0.0, 30.0)$  \\\hline
		\multirow{2}{*}{PEDOT:PSS}            & SLD ($\text{\AA}^{-2}$)  & 1.5               & $\mathcal{U}(0.0, 5.0)$   \\
		                                      & Thickness ($\text{\AA}$) & 600               & $\mathcal{U}(0.0, 800.0)$ \\\hline
		                                      & \rev{Interfacial roughness ($\text{\AA}$)}    & 5.0               & $\mathcal{U}(0.0, 8.0)$   \\\hline
		\multirow{2}{*}{Native oxide (Si02)}  & SLD ($\text{\AA}^{-2}$)  & 3.0               & $\mathcal{U}(0.0, 5.0)$   \\
		                                      & Thickness ($\text{\AA}$) & 50                & $\mathcal{U}(0.0, 800.0)$ \\\hline
		                                      & \rev{Interfacial roughness ($\text{\AA}$)}    & 5.0               & $\mathcal{U}(0.0, 8.0)$   \\\hline
		\multirow{2}{*}{Silicon backing (Si)} & SLD ($\text{\AA}^{-2}$)  & 2.07              & $\mathcal{U}(0.0, 5.0)$   \\
		                                      & Thickness ($\text{\AA}$) & $\infty$          &                           \\
	\end{tabular}
\end{table}

\begin{table}
	\centering
	\caption{NR parameters for the LED data including initial parameter guesses for optimisation and priors used for inference.}
	\label{tbl:LED_nrparam}
	\begin{tabular}{l|ll}
		Parameter                        & Initial value & Prior distribution                     \\\hline\hline
		Scale parameter (per device)     & 1.0           & $\mathcal{U}(0.9, 1.5)$                \\
		Background (shared)              & 5.0           & $\mathcal{U}(e^{-20}, 1\times10^{-7})$ \\
		Smearing parameter (\%) (shared) & 1.0           & $\mathcal{U}(0.0, 10.0)$               \\
	\end{tabular}
\end{table}

\begin{figure}
	\centering
	\begin{subfigure}{\fw}
		\includegraphics[width=\linewidth]{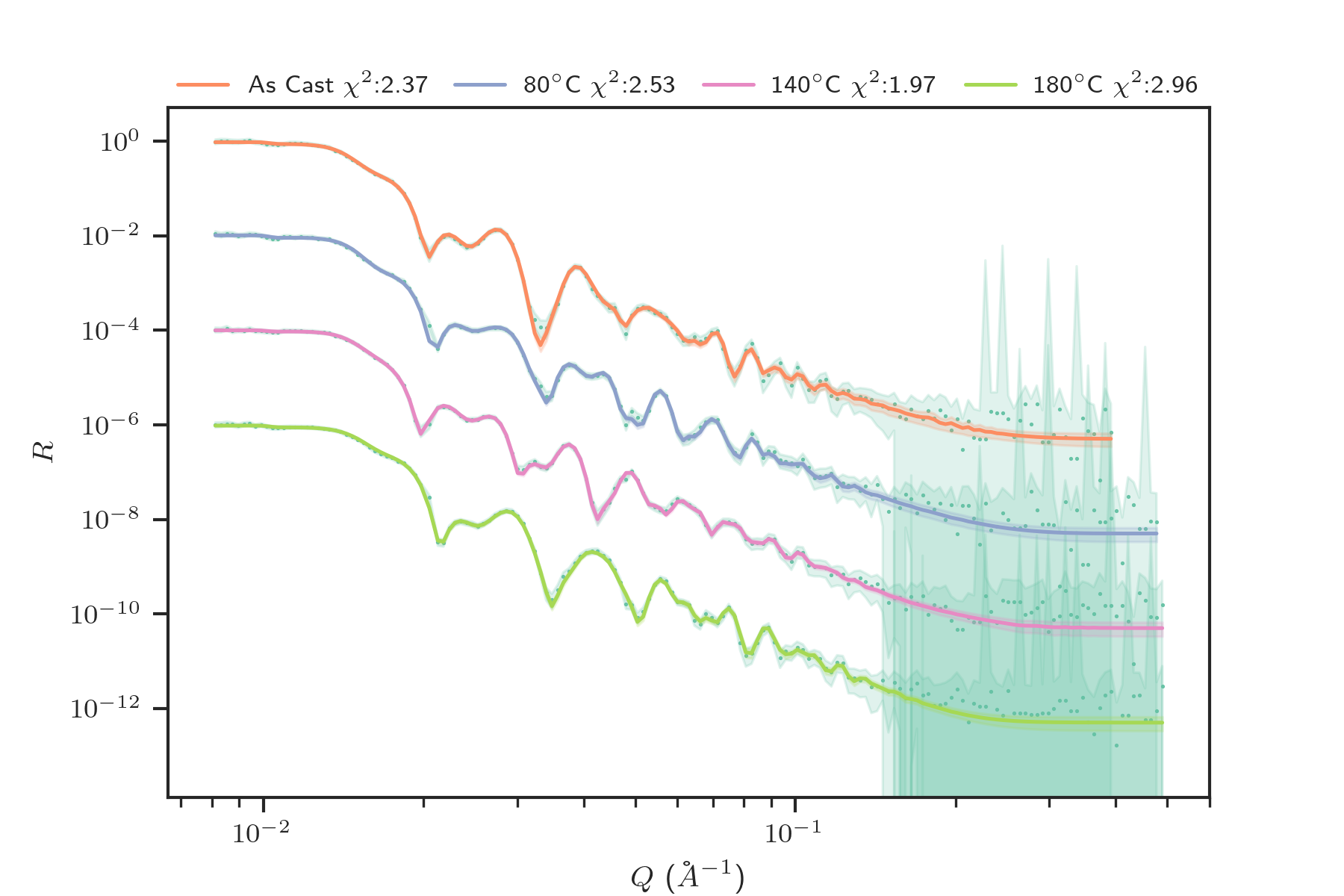}
		\caption{}
		\label{fig:sub1}
	\end{subfigure}
	%   \hfill
	\begin{subfigure}{\fw}
		\includegraphics[width=\linewidth]{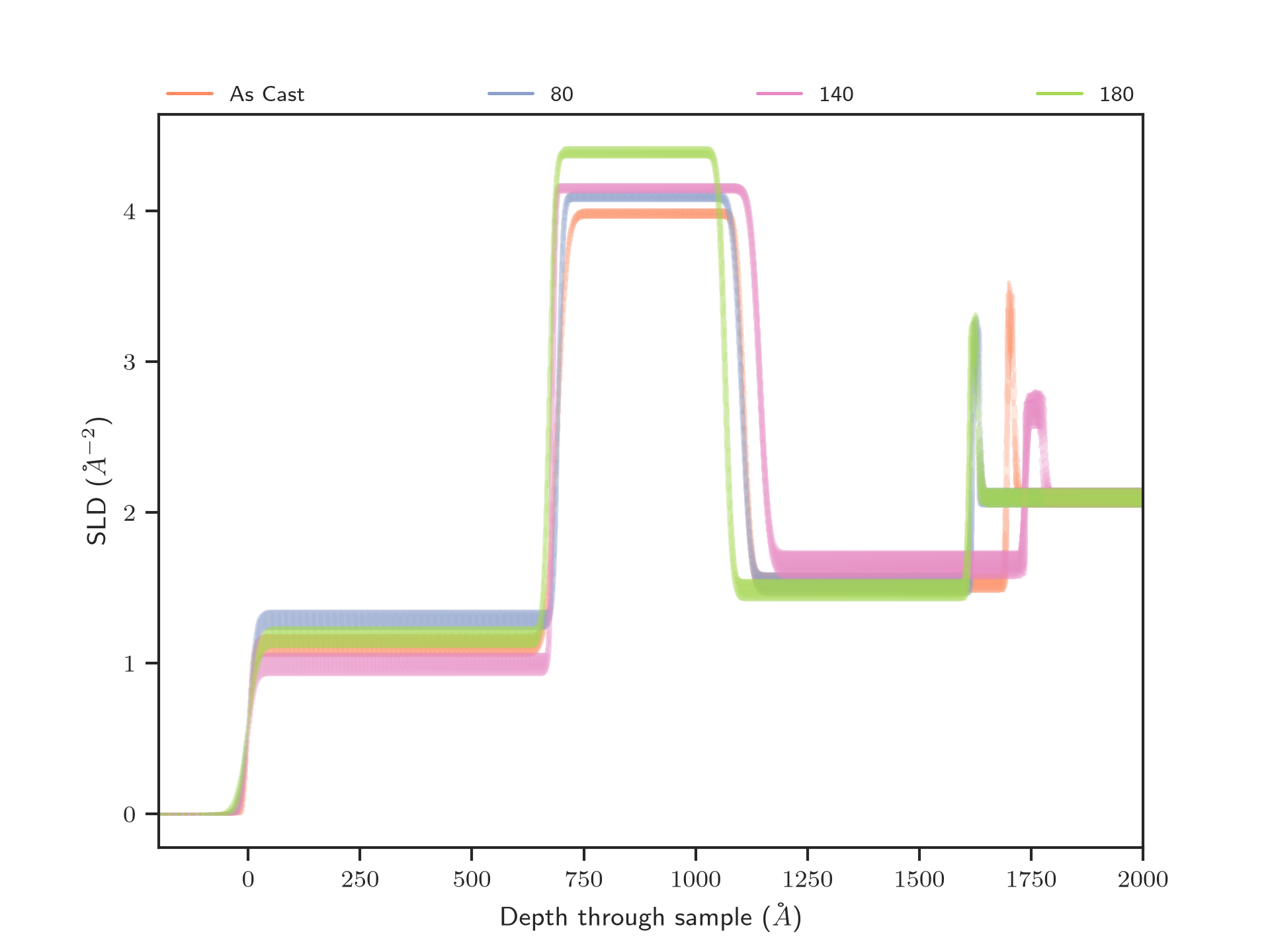}
		\caption{}
		\label{fig:sub2}
	\end{subfigure}
	\caption{(a) Results of the inference for the LED devices. Note that the reflectivity curves are spaced for visibility on the y-axis. (b) Samples from the distributions of the posterior SLD profiles.}
	\label{fig:LED}
\end{figure}

To produce an approximate inference over the co-fit posterior of all four devices a VI scheme is adopted. The surrogate posterior $q$ is selected to be a multivariate normal distribution defined in an unconstrained space. That is, all parameters are mapped from their physical bounded space to one that is unconstrained by an appropriate invertible transformation. The resulting ELBO is evaluated by automatic differentiation (using the numpyro library) and the refjax software for reflectivity kernel calculations. The mean of the surrogate posterior is initialised at identical values for each device and the variance is initialised as a diagonal matrix $\sigma_0I$ with $\sigma_0=0.1$. The ELBO is optimised using the ADAM \cite{kingma2014adam} optimiser for 2000 iterations with a learning rate of 0.05. To ensure that the optimisation does not become trapped in local minima, the optimisation procedure is run 100 times and the best (largest) value of the ELBO is retained as surrogate posterior. VI has significant computational advantages over sampling based approaches, each run of the inference is complete in <20 seconds on our hardware (AMD Ryzen Threadripper PRO 5995WX). Successive restarts can be run entirely in parallel and so the VI approach is very attractive with regard to modern CPU and GPU acceleration.

The predictive results of the inference are plotted in Figure \ref{fig:LED} \rrev{(for full results see Supplementary Information Section~S1)}. As can be seen in the Figure, the fit to data is very good with $\chi^2$ < 3 in all cases. Examining the SLD profiles, it appears that the samples annealed for the longest have higher SLD in the TFB layer due to extended annealing above the glass transition temperature $T_g$, this will help remove any trapped processing solvent as well as enable the layer to achieve an equilibrium density. We also see changes in the interfacial width between the two polyfluorenes which we are able to more rigorously quantify using this fitting approach. It is of note that the identified SLD of the Silicon backing ($2.09\pm0.0178$) is fit very close to the literature value of 2.07 indicating convergence to a physically plausible slab model. \rrev{The posterior distribution over the thickness of the native oxide layer for the device annealed at 140\,$^{\circ}$C is markedly larger than for the other three devices. Because the reflectivity is only very weakly sensitive to this thin, buried layer, we caution that this difference is almost certainly a fitting artifact rather than a genuine physical change between the devices, and the oxide-layer parameters should be interpreted with corresponding care. We further note that, for the as-cast device, the measured reflectivity displays slightly higher-amplitude fringes at high $Q$ than the fitted model (Figure \ref{fig:LED}a); we attribute this to the single resolution-smearing parameter that is shared across all four devices, which may slightly over-smear the as-cast curve. Device-specific sample non-uniformity or layer polydispersity, which are not captured by the shared smearing and slab model, may also contribute. A demonstration of both the HMC and VI approaches on a standard, previously published lipid bilayer dataset \cite{nelson2019refnx} is provided in Supplementary Information Section~S2, together with a discussion of the variance underprediction inherent to VI in Supplementary Information Section~S3.} 

\section{Discussion}

The results of our work demonstrate the potential of automatic differentiation and gradient-based optimisation techniques in the context of neutron reflectometry data analysis. Excellent fits to both simple and complex NR data are demonstrated, in one case exceeding the quality of fit compared to previous literature results despite wide priors.

In the optimisation setting, access to cutting edge gradient descent algorithms will enable practitioners to quickly and efficiently fit NR data, particularly as samples become more complex and traditional approaches (LM, BFGS) become prone to computational complexity and become stuck in local minima in the large data limit.

In the inference setting, the application of Hamiltonian Monte Carlo to NR data has shown considerable sample efficiency gains over traditional gradient-free MCMC methods. HMC is a promising replacement for MCMC in NR in scenarios where accurate uncertainty quantification is required and complex posterior distribution geometries (multimodality, highly correlated variables) make NR inference challenging.

In the second case-study, techniques from variational inference show remarkable performance in a complex NR fitting task. \rev{VI is shown to be a potentially revolutionary technique for several NR applications} particularly in settings whereby uncertainty quantification of high-throughput data is critical. An example is fast kinetic NR experiments whereby large quantities of NR data are collected dynamically and in-situ inference of model parameters is required. A key application of VI in NR might be in real-time parameter inference for closed-loop process control, something that has thus far required fast model surrogates and has been unattainable with traditional inverse modelling techniques. Another application could be in iterative experimental design whereby the parameter posterior distributions of initial experiments are used to inform the design of subsequent experiments during a single beamtime. Here, techniques from Bayesian optimisation \cite{ru2018fast} may provide a principled way to select the most informative experiments sequentially. A related opportunity is to use the proposed approach to provide fast inversion to be used during active learning of sequential NR experiments as in \cite{hoogerheide2024autorefl}.

Alongside the case-studies developed here, the authors make available an open-source software library for fast computation of forward reflectivity and gradients in the python language. \rev{It is the intention of the authors to use this platform as a springboard to develop further forward reflectometry models for other indirect measurement techniques such as small-angle neutron scattering, ellipsometry and grazing-incidence X-ray scattering (where HMC has previously been shown to be a successful approach \cite{zhang2022parameter}).}
In conclusion, \rev{we have demonstrated that the availability of fast-forward reflectometry models is critical to ensure that NR fitting keeps pace with high-throughput experimentation} and the rapid advances in ML and optimisation techniques. This paper presents a promising step in this direction by ensuring that gradient-based inverse methods are readily available for NR data analysis.

\section*{Author contributions}

MDC: Conceptualization, Data curation, Writing - Original Draft, Visualization, Software; AJP: Conceptualization, Data curation, Writing - Review \& Editing, Funding acquisition; PG: Data curation, Writing - Review \& Editing; MAS: Writing - Review \& Editing; JPAF: Funding acquisition, Writing - Review \& Editing; TJR: Conceptualization, Writing - Review \& Editing, Funding acquisition; SLB: Conceptualization, Funding acquisition, Writing - Review \& Editing;

\section*{Funding statement}

SB, PF and AJP were supported via an EPSRC prosperity partnership, called Sustainable Coatings by Rational Design, SusCoRD, EP/S004963/1, this was jointly funded by the company AkzoNobel.

The OLED data was collected on the ISIS neutron reflectometer instrument SURF, and we are grateful to ISIS for beamtime at ISIS under beamtime application RB 1120359 (DOI 10.5286/ISIS.E.24088355).  We are grateful for their support and encouragement of this work. 

\section*{Competing interests}

We confirm that there are no competing interests to declare. 

\section*{Acknowledgements}

We acknowledge Dr Hiroshi Hamamatsu from Sumitomo Chemicals Corporation for producing the OLED device samples whilst on secondment to the University of Sheffield. We are also indebted to Cambridge Display Technologies (CDT) for synthesising and supplying the deuterated conjugated polymers used in the OLED devices. This work was (partially) supported by the Centre for Machine Intelligence at the University of Sheffield.

\bibliographystyle{unsrtnat}
\bibliography{refjax.bib}

\end{document}

\maketitle

\section{Posterior distributions}

\begin{figure}
	\centering
	\includegraphics[width=\linewidth]{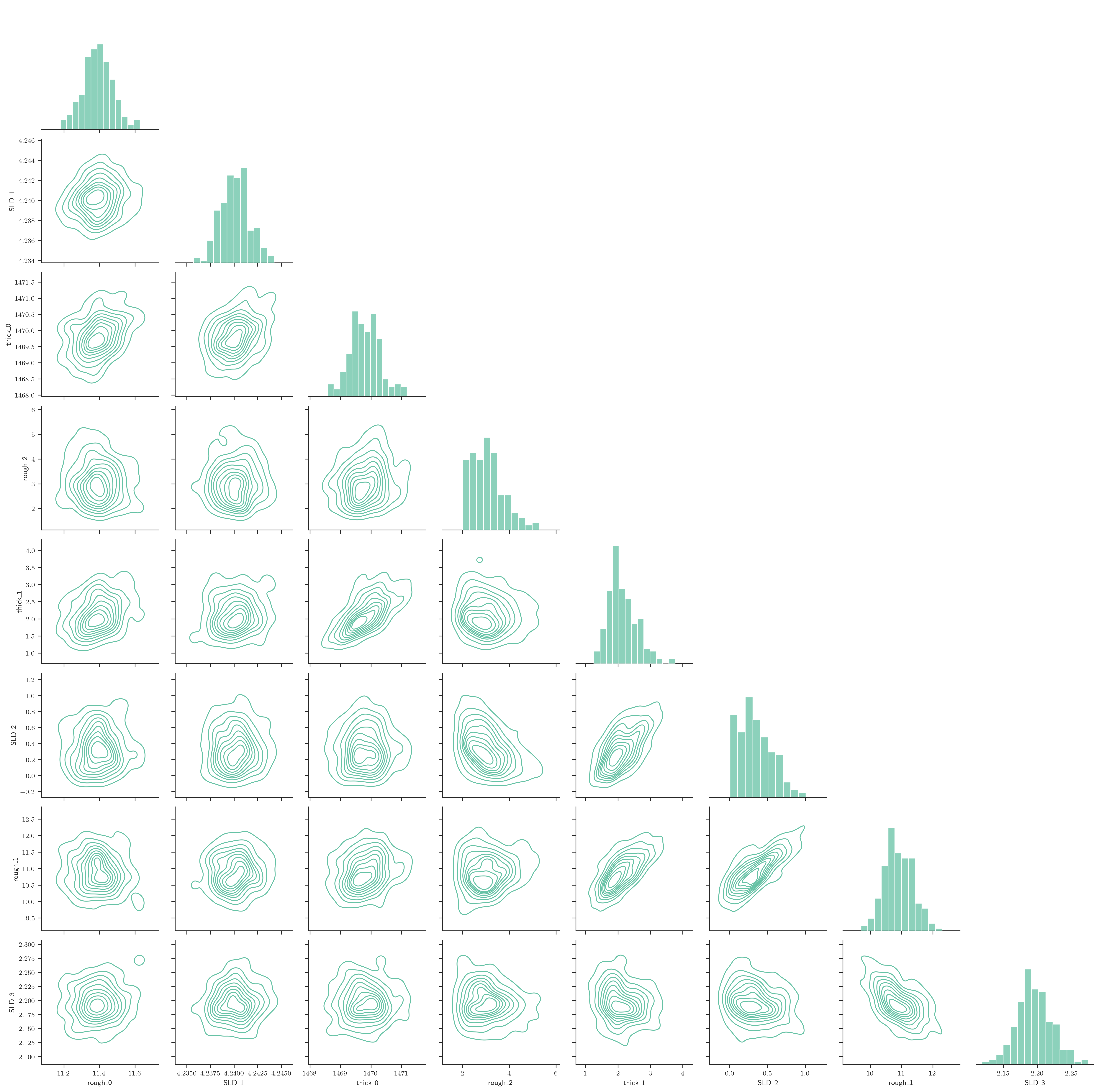}
	\caption{Posterior distribution identified over the slab model parameters of the crystalline quartz sample.}
	\label{fig:q_post}
\end{figure}

\begin{figure}
	\centering
	\includegraphics[width=\linewidth]{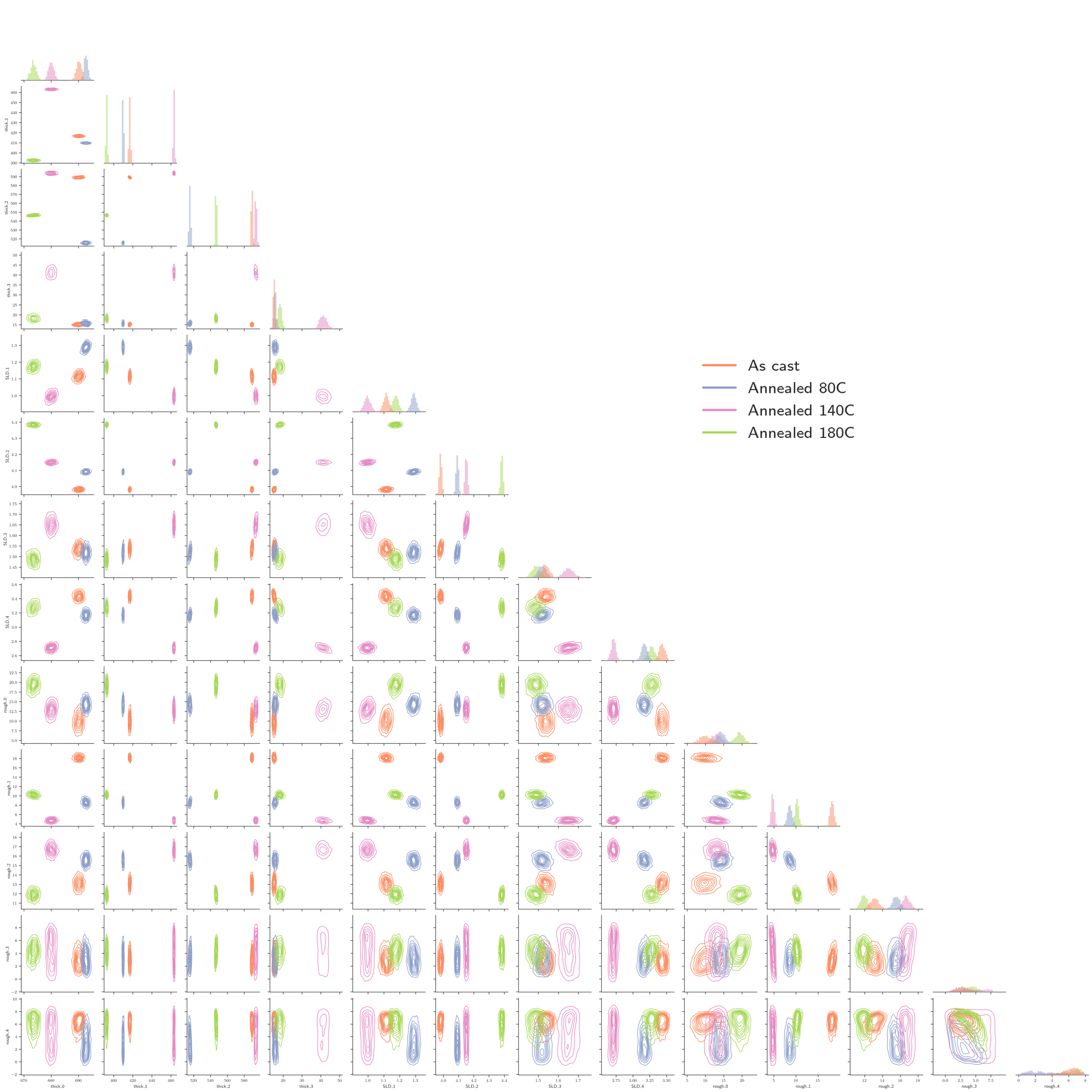}
	\caption{Posterior distribution identified over the slab model parameters of the LED devices.}
	\label{fig:led_post}
\end{figure}

\begin{figure}
	\centering
	\begin{subfigure}{0.49\linewidth}
		\includegraphics[width=\linewidth]{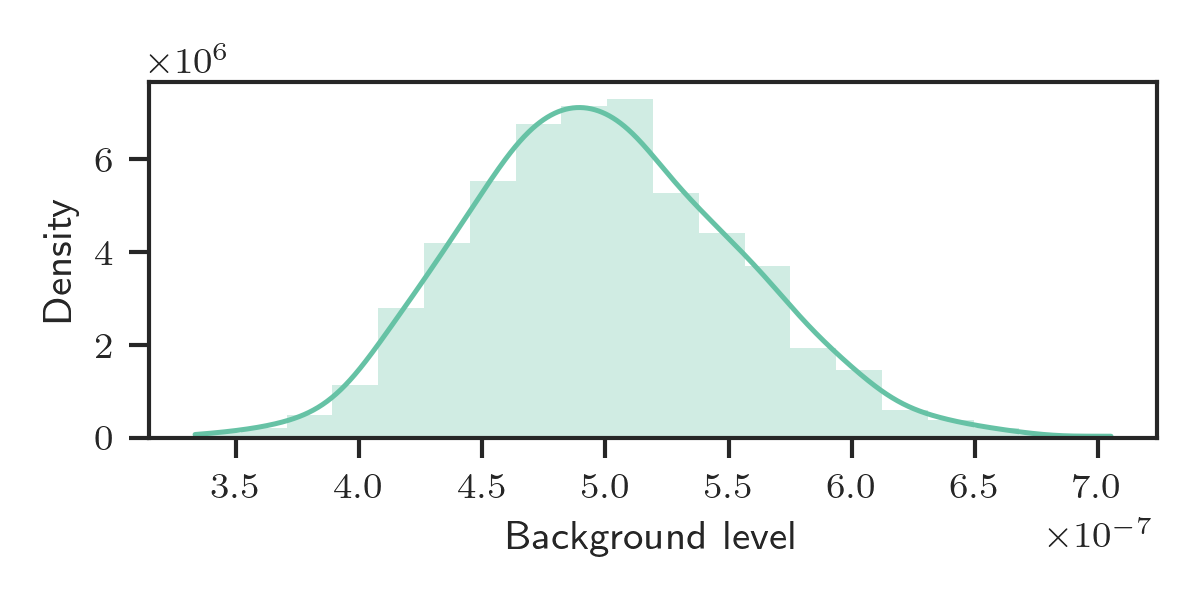}
		\caption{}
		\label{fig:ledsub1}
	\end{subfigure}
	\begin{subfigure}{0.49\linewidth}
		\includegraphics[width=\linewidth]{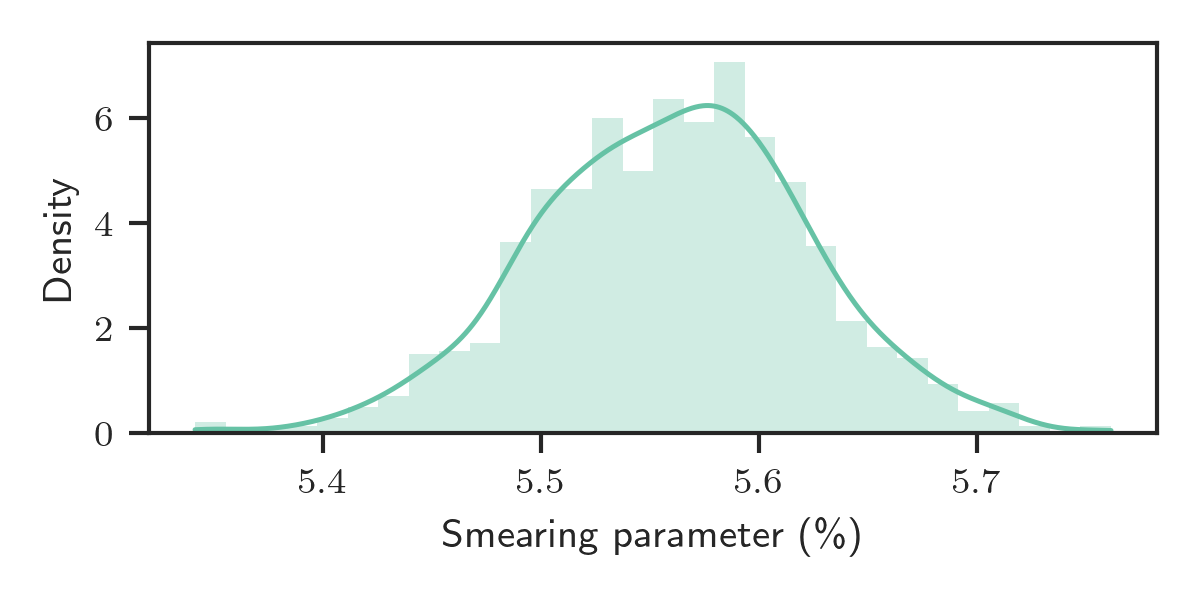}
		\caption{}
		\label{fig:ledsub2}
	\end{subfigure}
	\begin{subfigure}{\fw}
		\includegraphics[width=\linewidth]{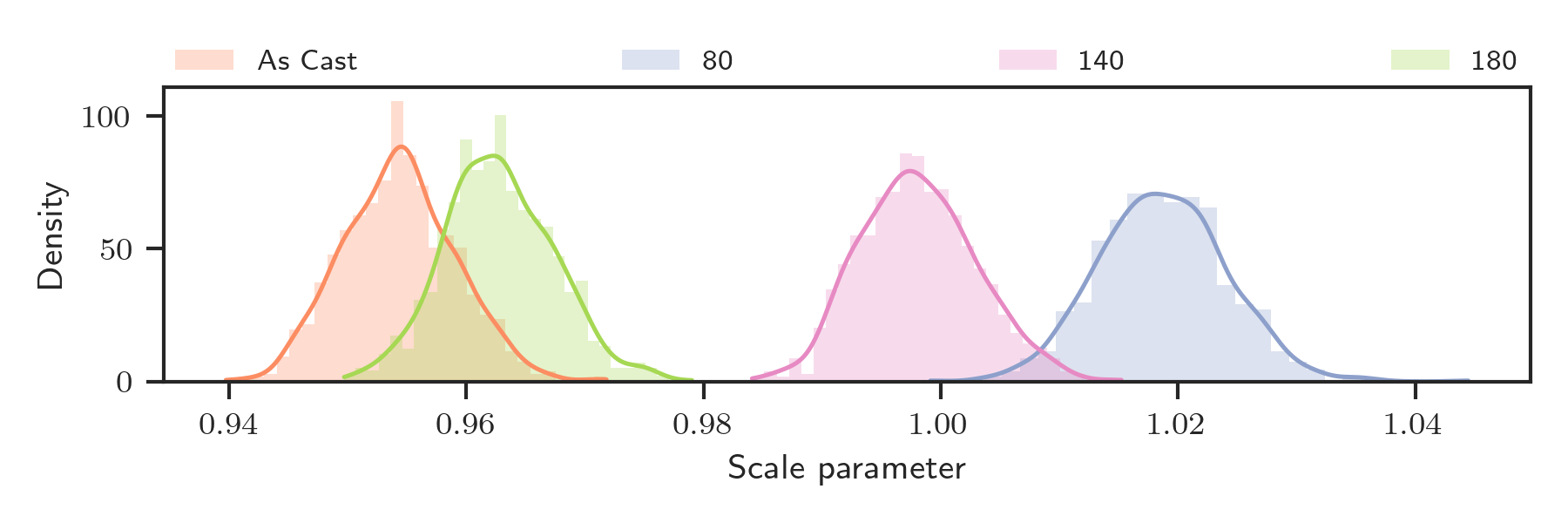}
		\caption{}
		\label{fig:ledsub3}
	\end{subfigure}
	\caption{Identified VI posterior distributions of NR parameters for the LED data. (a) Background level (shared between all devices). (b) Smearing parameter (shared between all devices). (c) Scale parameter.}
	\label{fig:LED_app}
\end{figure}

Posterior distributions identified in this study are included here. In order to limit the size of the figures, only the distributions of the slab model parameters are included in the corner plots. Posterior distributions for the quartz and LED data are plotted in Figures \ref{fig:q_post} and \ref{fig:led_post} respectively. Distributions of the instrument parameters are plotted in Figure \ref{fig:LED_app}. Posterior parameter values from the LED data are collected in Table \ref{tbl:LED_post} for the slab model parameters and in Table \ref{tbl:LED_nrparam_post} for the instrument parameters.

\begin{table}[h]
	\centering
	\caption{Identified posterior means and variances from the LED device data. Parameters without values are taken to be fixed at their initial values given in Table 3 of the main manuscript.}
	\label{tbl:LED_post}
	\begin{tabular}{l|llllll}
		Layer                      & Parameter                & As Cast          & 80C             & 140C             & 180C             \\\hline\hline
		\multirow{2}{*}{Air}       & SLD ($\text{\AA}^{-2}$)  &                                                                          \\
		                           & Thickness ($\text{\AA}$) &                                                                          \\\hline
		                           & Roughness                & $10.1\pm1.68$    & $14.3\pm1.14$   & $13\pm1.36$      & $19.3\pm1.24$    \\\hline
		\multirow{2}{*}{F8BT}      & SLD ($\text{\AA}^{-2}$)  & $1.11\pm0.0184$  & $1.29\pm0.017$  & $0.996\pm0.0212$ & $1.17\pm0.0189$  \\
		                           & Thickness ($\text{\AA}$) & $690\pm0.933$    & $693\pm0.721$   & $680\pm1$        & $673\pm1.06$     \\\hline
		                           & Roughness                & $18.1\pm0.443$   & $8.58\pm0.546$  & $4.74\pm0.335$   & $10.2\pm0.41$    \\\hline
		\multirow{2}{*}{TFB}       & SLD ($\text{\AA}^{-2}$)  & $3.98\pm0.00779$ & $4.09\pm0.0079$ & $4.15\pm0.00827$ & $4.38\pm0.00777$ \\
		                           & Thickness ($\text{\AA}$) & $417\pm0.716$    & $410\pm0.513$   & $463\pm0.591$    & $393\pm0.67$     \\\hline
		                           & Roughness                & $13.1\pm0.49$    & $15.5\pm0.411$  & $16.7\pm0.458$   & $11.9\pm0.387$   \\\hline
		\multirow{2}{*}{PEDOT:PSS} & SLD ($\text{\AA}^{-2}$)  & $1.54\pm0.0188$  & $1.52\pm0.0203$ & $1.65\pm0.0263$  & $1.49\pm0.0205$  \\
		                           & Thickness ($\text{\AA}$) & $589\pm0.786$    & $516\pm0.934$   & $594\pm1.11$     & $547\pm0.79$     \\\hline
		                           & Roughness                & $2.89\pm1.09$    & $3.15\pm1.31$   & $4.44\pm2.17$    & $4.26\pm1.09$    \\\hline
		\multirow{2}{*}{Si02}      & SLD ($\text{\AA}^{-2}$)  & $3.43\pm0.0429$  & $3.17\pm0.0431$ & $2.71\pm0.0337$  & $3.28\pm0.0546$  \\
		                           & Thickness ($\text{\AA}$) & $15.1\pm0.451$   & $15.7\pm0.646$  & $41.3\pm2.02$    & $18.2\pm1.02$    \\\hline
		                           & Roughness                & $6.22\pm0.895$   & $2.89\pm1.71$   & $4.46\pm2.27$    & $6.08\pm1.28$    \\\hline
		\multirow{2}{*}{Si}        & SLD ($\text{\AA}^{-2}$)  & $2.09\pm0.0178$  & $2.09\pm0.0178$ & $2.09\pm0.0178$  & $2.09\pm0.0178$  \\
		                           & Thickness ($\text{\AA}$) &                                                                          \\
	\end{tabular}
\end{table}

\begin{table}
	\centering
	\caption{Identified means and variances of the NR parameters from the LED devices.}
	\label{tbl:LED_nrparam_post}
	\begin{tabular}{l|l}
		Parameter                 & Value                \\\hline\hline
		Background                & $4.99e-07\pm5.5e-08$ \\
		Smearing parameter (\%)   & $5.56\pm0.0624$      \\
		Scale parameter (As cast) & $0.954\pm0.00472$    \\
		Scale parameter (80C)     & $1.02\pm0.00542$     \\
		Scale parameter (140C)    & $0.998\pm0.00489$    \\
		Scale parameter (180C)    & $0.963\pm0.00465$    \\
	\end{tabular}
\end{table}

\section{Benchmark performance on a lipid bilayer system}

This section presents results on a benchmark bilayer lipid system that has been previously studied in other published software for the analysis of NR data \cite{nelson2019refnx}. As such, this system makes an ideal benchmark for comparison of new fitting methodologies.

\begin{figure}
    \centering
    \includegraphics[width=0.8\linewidth]{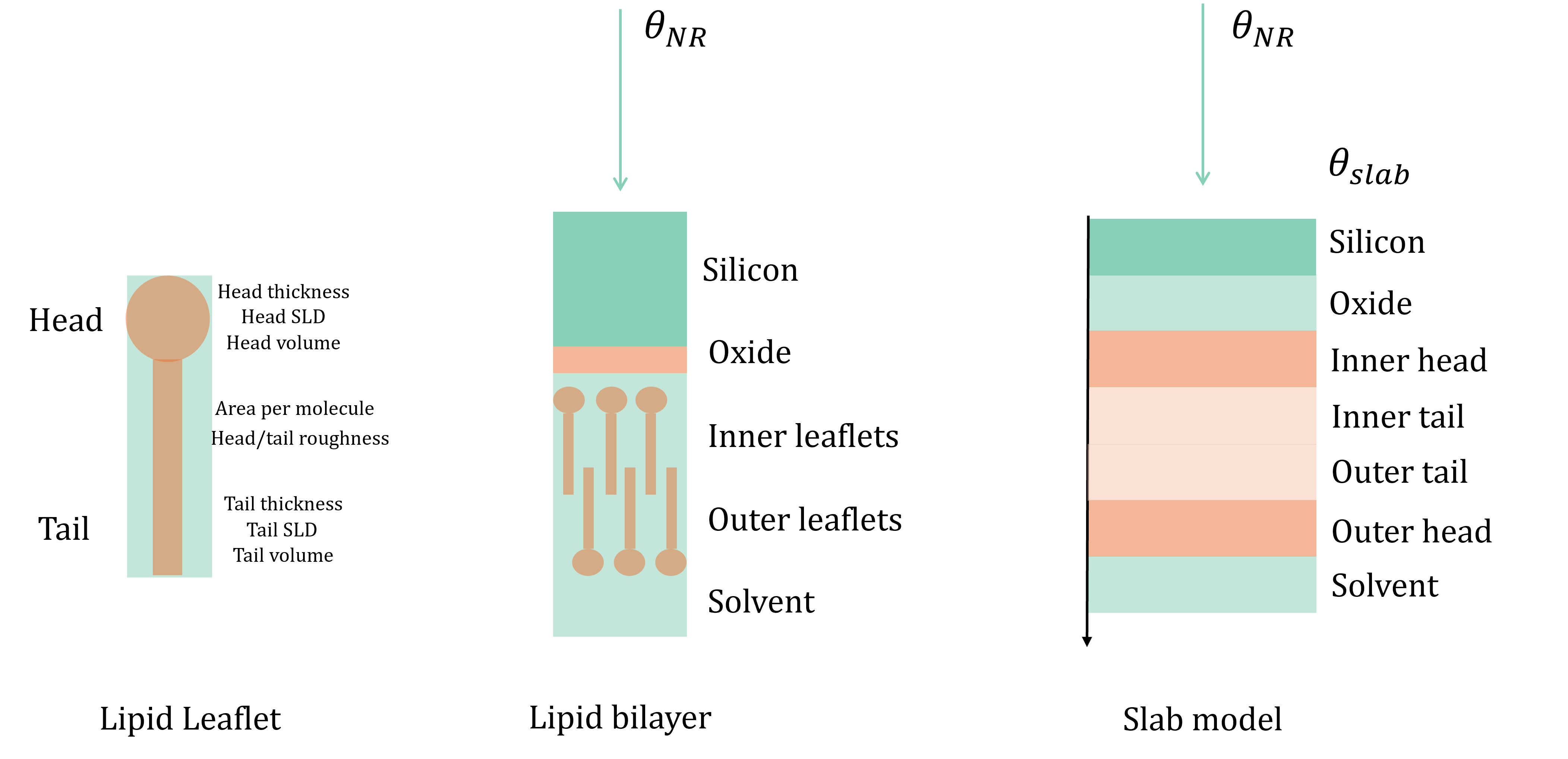}
    \caption{Slab model of a bilayer Lipid. Each slab on the right is characterised by a single SLD and thickness. Each interface is characterised by a Gaussian roughness.}
    \label{fig:lip_schem}
\end{figure}

The data comprises three NR curves from a lipid bilayer system with different solvents. The solvents considered are D2O, H2O and mixed solvent of D2O and H2O. Figure \ref{fig:lip_schem} depicts the slab model that is used both in \cite{nelson2019refnx} and here to represent the system.

The slab model parameters are derived from the properties of the inner and outer lipid leaflets, the silicon backing, the oxide layer and the solvent. The details of this parameterisation are identical to those in \cite{nelson2019refnx} and so are omitted here. Overall there are 78 parameters in the model, however many of these are assumed to be known in the analysis. The authors note that in the approach presented here, all three solvent SLDs are treated as unknown and a different background and scale parameter are inferred for each solvent. In total this approach yields 17 free parameters (6 corresponding to instrument settings and 11 corresponding to the slab model) rather than the 13 free parameters considered in \cite{nelson2019refnx}.

In order to compare inference approaches (and also to demonstrate the effects of variance underestimation in VI) inference will be attempted with both approaches form the paper. HMC with the No U-turn sampler, and VI by gradient descent.

\begin{figure}
	\centering
	\begin{subfigure}{0.8\linewidth}
		\includegraphics[width=\linewidth]{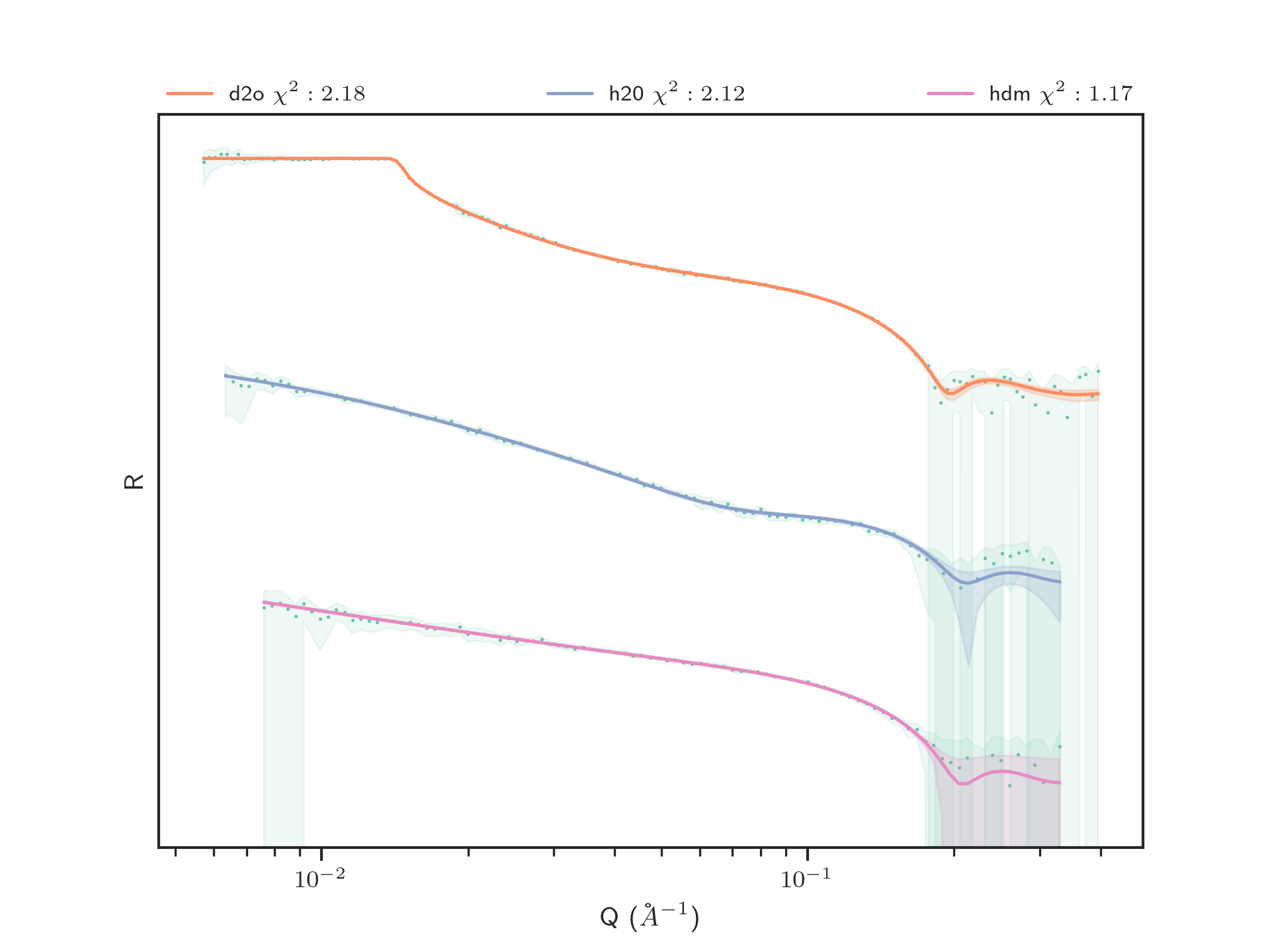}
		\caption{VI}
		\label{fig:lip_sub11}
	\end{subfigure}
	%   \hfill
	\begin{subfigure}{0.8\linewidth}
		\includegraphics[width=\linewidth]{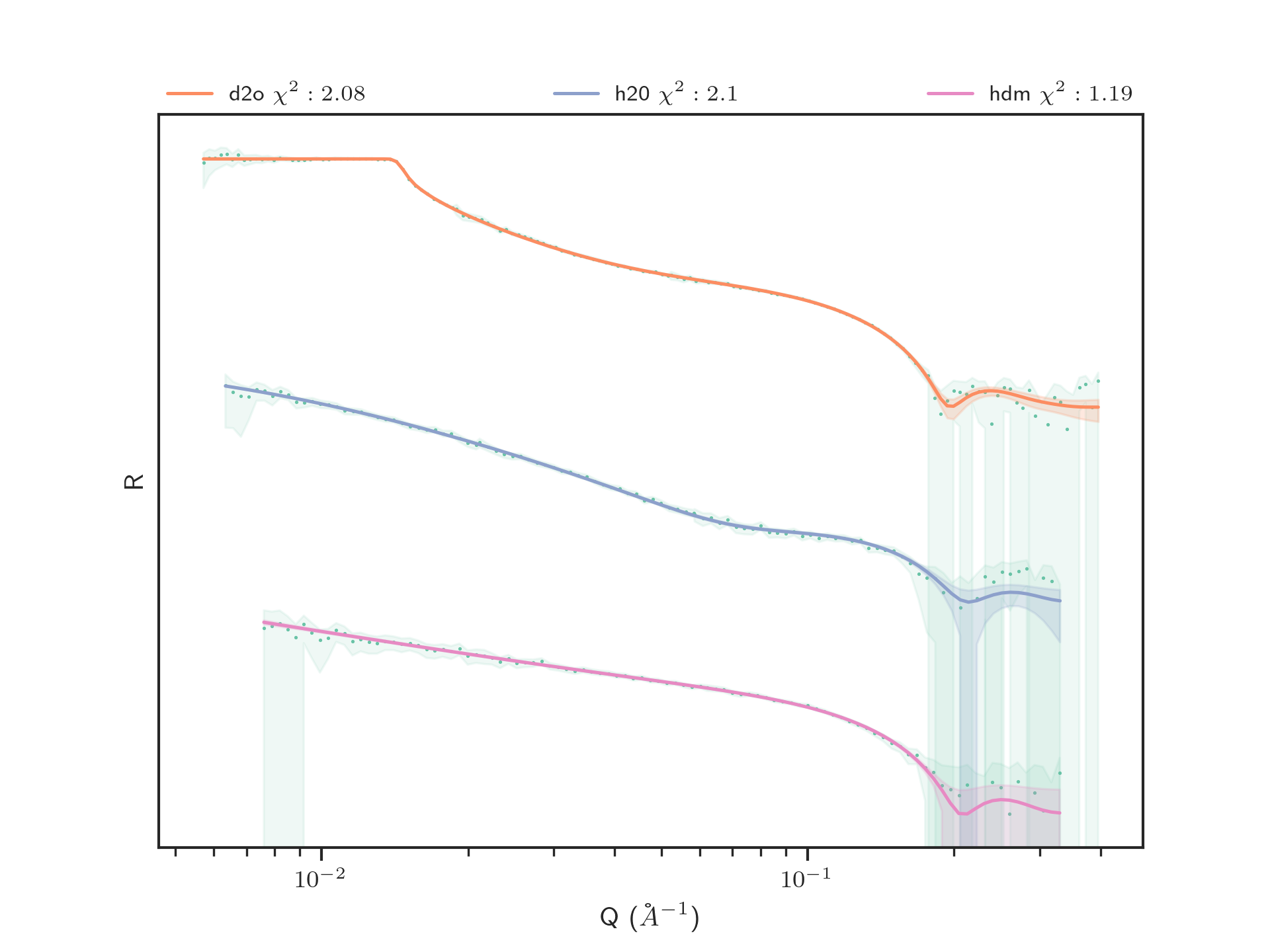}
		\caption{HMC}
		\label{fig:lip_sub12}
	\end{subfigure}
	\caption{Results of the inference for the lipid bilayer -- Predicted reflectometry curves. (a) Posterior estimated by a variational inference scheme with a Gaussian surrogate on the unconstrained space. (b) posterior estimated by Hamiltonian Monte-Carlo Note that the reflectivity curves are spaced for visibility on the y-axis.}
	\label{fig:lip_pred}
\end{figure}

\begin{figure}
	\centering
	\begin{subfigure}{0.8\linewidth}
		\includegraphics[width=\linewidth]{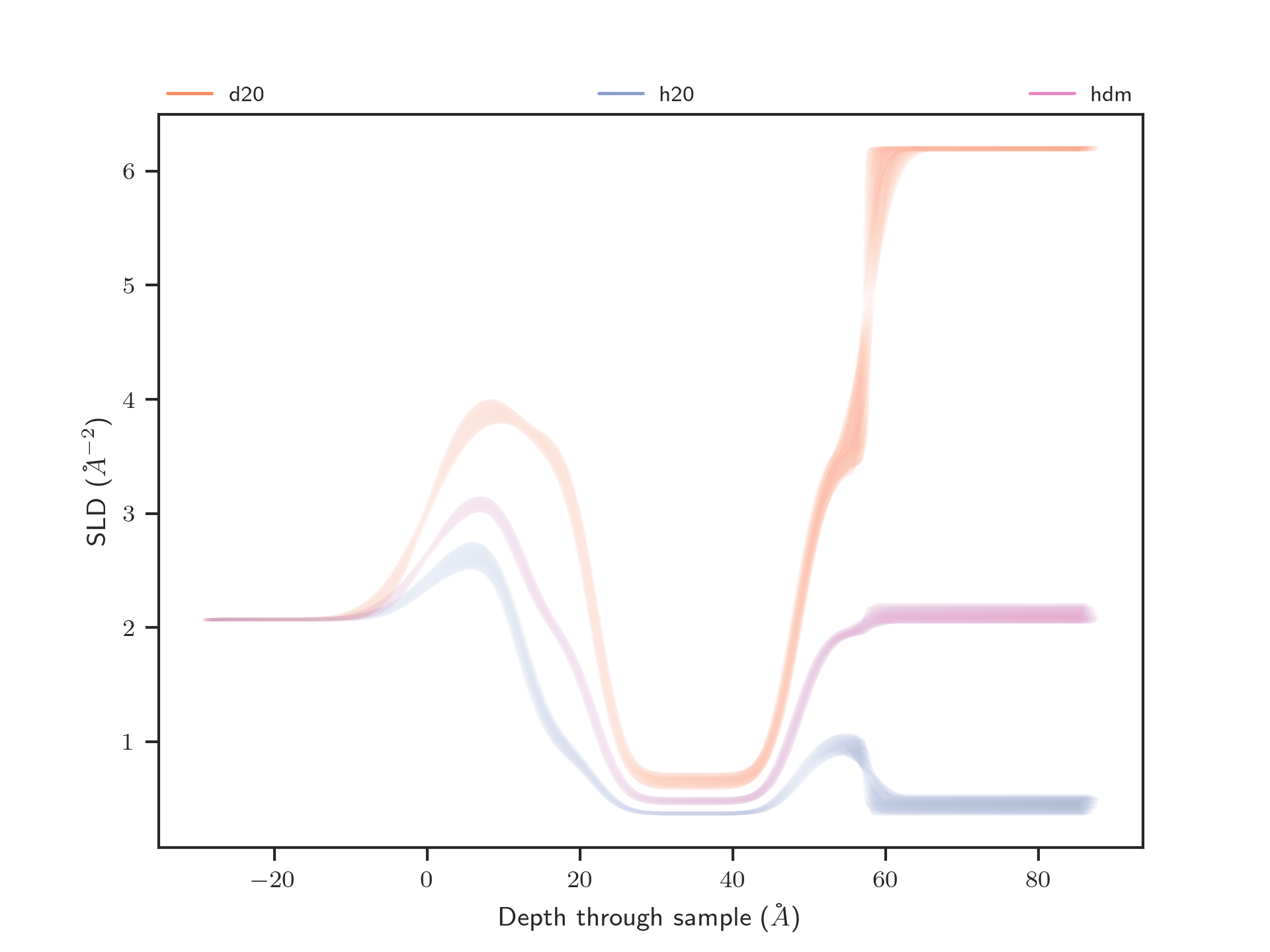}
		\caption{VI}
		\label{fig:lip_sub21}
	\end{subfigure}
	%   \hfill
	\begin{subfigure}{0.8\linewidth}
		\includegraphics[width=\linewidth]{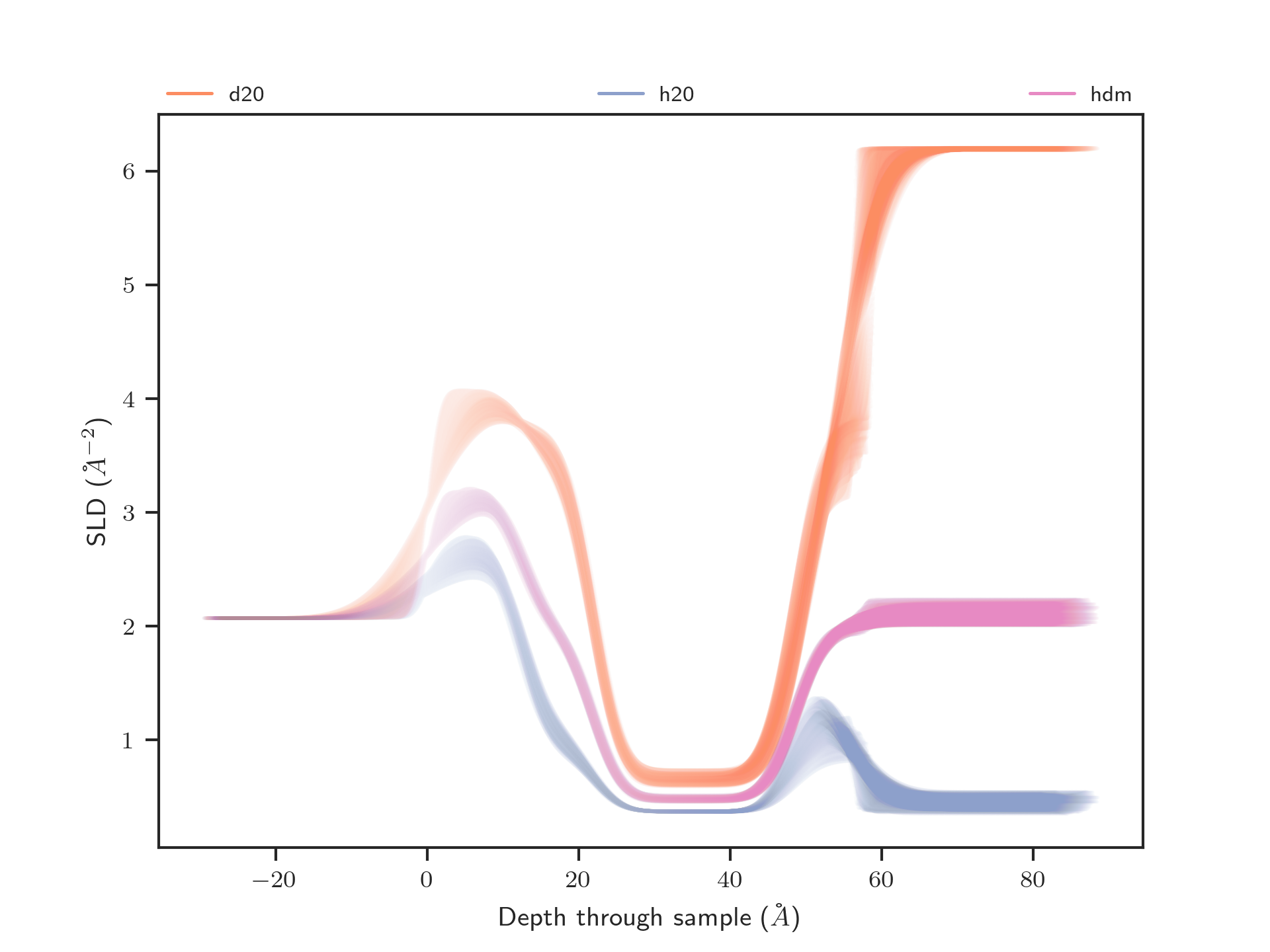}
		\caption{HMC}
		\label{fig:liip_sub22}
	\end{subfigure}
	\caption{Results of the inference for the lipid bilayer -- SLD profiles. (a) Posterior estimated by a variational inference scheme with a Gaussian surrogate on the unconstrained space. (b) posterior estimated by Hamiltonian Monte-Carlo Note that the reflectivity curves are spaced for visibility on the y-axis.}
	\label{fig:lip_prof}
\end{figure}

\begin{figure}
	\centering
	\includegraphics[width=\linewidth]{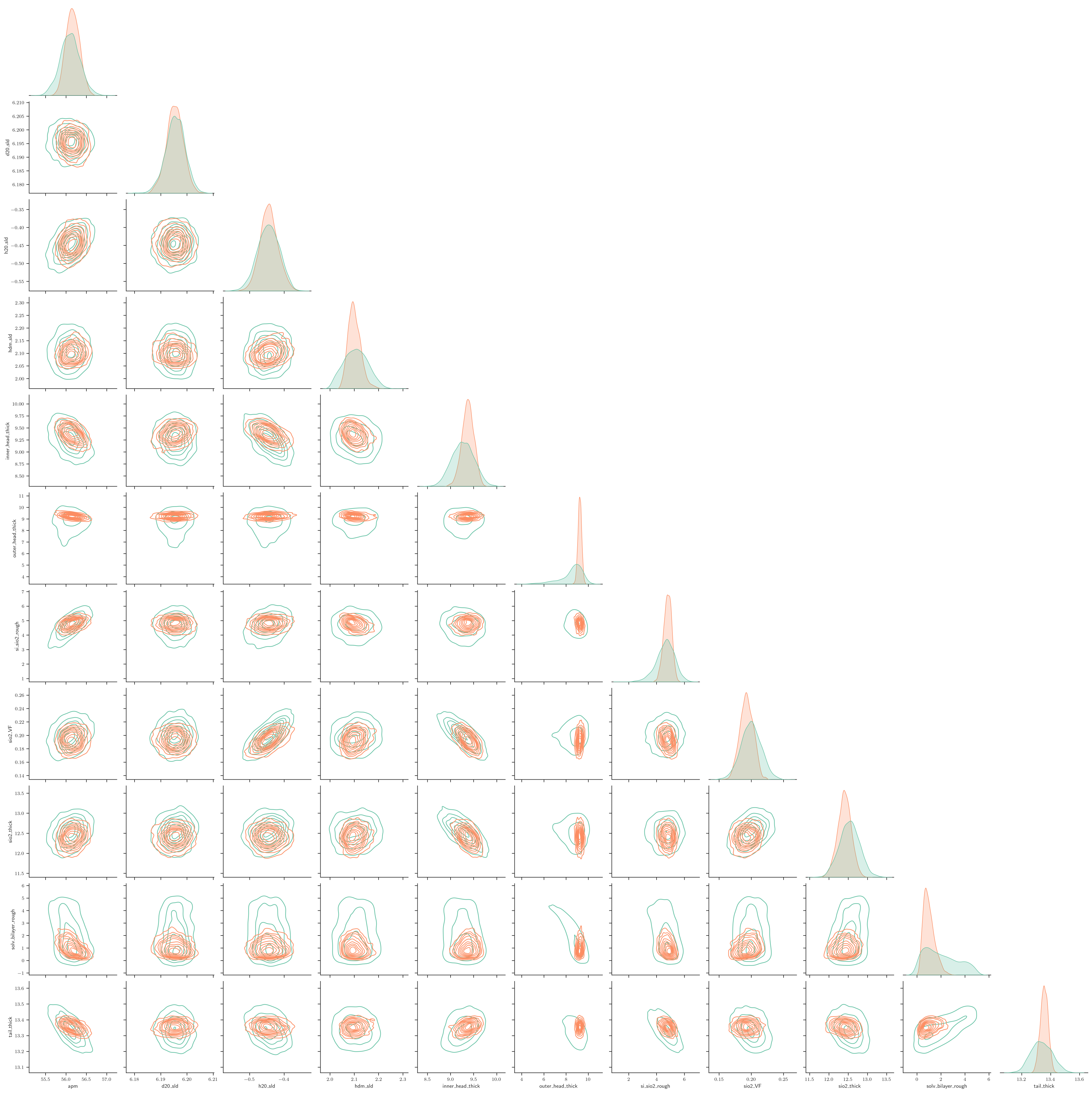}
	\caption{Posterior corner plot of identified parameters from both the NUTS (green) and VI (orange) inference approaches. Note the reduced variance in the VI posterior, particularly evident when the posterior is non-Gaussian. Note that for reasons of space, only parameters corresponding to the slab model are plotted here.}
	\label{fig:lip_corner_slab}
\end{figure}

As can be seen in Figure \ref{fig:lip_pred} the fit to the data is very strong in both cases ($\chi^2 < 3$ in all cases). It can also be seen in Figures \ref{fig:lip_prof} and \ref{fig:lip_corner_slab} that the posterior variance is slightly lower in the VI result. This is not unexpected and is a known tendency of VI approaches to be `mode-seeking'.

\newpage
\section{Remark on the underprediction of variance in VI}

As can be seen in the figures of the previous section, the fit to the data is very strong in both cases ($\chi^2 < 3$ in all cases). We also see in the results that the posterior variance is slightly lower in the VI results. This is not unexpected and is a known property of VI approaches.

The reason for the variance contraction can be attributed to the KL divergence used to measure the distance between the surrogate and the true posterior.

\begin{equation}
	\operatorname{KL}(q||p) = \int q(\theta) \log \frac{q(\theta)}{p(\theta)}d\theta
\end{equation}

Where the surrogate distribution $q(z)$ is large and the true posterior $p(z)$ is small this will result in large divergences and punish models that incorrectly assign probability mass where it should not (high-probability regions get emphasised). In contrast, When $q(z)$ is small then there is almost no contribution to the divergence. The result is that the divergence does not punish models that miss modes or heavy tails of a distribution (low-probability tails are underrepresented).

Overall this results in the 'mode-seeking' behaviour of VI that can be seen in Figure \ref{fig:lip_corner_slab}. In areas where the posterior identified by HMC is non-Gaussian we see that the VI can miss the tails of these distributions despite forming a good approximation to the principal modes. This effect is compounded by the Gaussian surrogate assumption that is insufficiently flexible to account for nonlinear correlations or heavily skewed tails.

Although this variance underprediction is inconvenient, it is sometimes worth the massive computational efficiency savings to use VI over a sampling-based scheme. Ultimately, this decision will come down to how the propagated uncertainty is to be used. If one is only intending to propagate mean and variances of a few parameters through to further analyses (for example fitting a model to a changing parameter in a kinetic run) then VI may well be sufficient. However, for full characterisation of unknown systems it may be more appropriate to use a 'full-fat' sampling approach.

% Guard against a spurious "missing \item" error at the bibliography that can
% arise when floats are placed by the output routine with \if@newlist still set.
\makeatletter\@newlistfalse\makeatother
\bibliographystyle{unsrtnat}
\bibliography{refjax.bib}